\documentclass[10pt,twocolumn,letterpaper]{article}

\usepackage[pagenumbers]{cvpr}
\usepackage{graphicx}
\usepackage{amsmath}
\usepackage{amssymb}
\usepackage{booktabs}
\usepackage{pifont}
\usepackage[table,dvipsnames,math]{xcolor}
\usepackage{wrapfig}
\usepackage{tabularx}
\usepackage{tikz}
\usepackage{bbm}
\usepackage{multirow}
\usepackage{fontawesome}
\usepackage{marvosym}

\definecolor{lblue}{RGB}{66, 133, 244}
\definecolor{mmdet3d}{RGB}{1, 121, 194}

\usepackage[pagebackref=false,breaklinks,colorlinks,citecolor=lblue,urlcolor=lblue,linkcolor=lblue]{hyperref}

\usepackage[capitalize]{cleveref}
\crefname{section}{Sec.}{Secs.}
\Crefname{section}{Section}{Sections}
\Crefname{table}{Table}{Tables}
\crefname{table}{Tab.}{Tabs.}

\def\cvprPaperID{2023}
\def\confName{Technical Report}
\def\confYear{2024}

\begin{document}

\title{An Empirical Study of Training State-of-the-Art LiDAR Segmentation Models}

\author{
    Jiahao Sun$^{\heartsuit,1,2}$ \quad Chunmei Qing$^{\heartsuit,1,\textrm{\Letter}}$ \quad Xiang Xu$^{\heartsuit,3}$ \quad Lingdong Kong$^{\heartsuit,2,4}$ \quad Youquan Liu$^{\heartsuit,5}$\\
    Li Li$^{\heartsuit,6}$ \quad
    Chenming Zhu$^{\heartsuit,1,7}$ \quad Jingwei Zhang$^{\heartsuit,2}$ \quad Zeqi Xiao$^{\heartsuit,8}$ \quad Runnan Chen$^{\heartsuit,7}$
    \\
    Tai Wang$^{\heartsuit,2}$ \quad Wenwei Zhang$^{\heartsuit,2}$ \quad Kai Chen$^{\heartsuit,2}$
    \\\\
    $^{\mathbf{\heartsuit}}$MMDetection3D Contributors\\
    $^{1}$South China University of Technology
    \quad
    $^{2}$Shanghai AI Laboratory
    \\
    $^{3}$Nanjing University of Aeronautics and Astronautics
    \quad
    $^{4}$National University of Singapore
    \\
    $^{5}$Fudan University
    \quad
    $^{6}$Durham University
    \quad
    $^{7}$University of Hong Kong
    \\
    $^{8}$Nanyang Technological University, Singapore
    \\
    \url{https://github.com/open-mmlab/mmdetection3d}
}
\maketitle

\newcommand\blfootnote[1]{%
\begingroup
\renewcommand\thefootnote{}{}\footnote{#1}%
\addtocounter{footnote}{-1}%
\endgroup
}

\blfootnote{\textrm{\Letter} Corresponding Author: {qchm@scut.edu.cn}.}

%%%%%%%%% ABSTRACT
\begin{abstract}
   In the rapidly evolving field of autonomous driving, precise segmentation of LiDAR data is crucial for understanding complex 3D environments. Traditional approaches often rely on disparate, standalone codebases, hindering unified advancements and fair benchmarking across models. To address these challenges, we introduce \textcolor{lblue}{\textbf{MMDetection3D-lidarseg}}, a comprehensive toolbox designed for the efficient training and evaluation of state-of-the-art LiDAR segmentation models. We support a wide range of segmentation models and integrate advanced data augmentation techniques to enhance robustness and generalization. Additionally, the toolbox provides support for multiple leading sparse convolution backends, optimizing computational efficiency and performance. By fostering a unified framework, MMDetection3D-lidarseg streamlines development and benchmarking, setting new standards for research and application. Our extensive benchmark experiments on widely-used datasets demonstrate the effectiveness of the toolbox. The codebase and trained models have been publicly available, promoting further research and innovation in the field of LiDAR segmentation for autonomous driving.
\end{abstract}

%%%%%%%%% BODY TEXT
\section{Introduction}
\label{sec:intro}

\begin{figure}[t]
    \centering
    \includegraphics[width=\linewidth,scale=1.00]{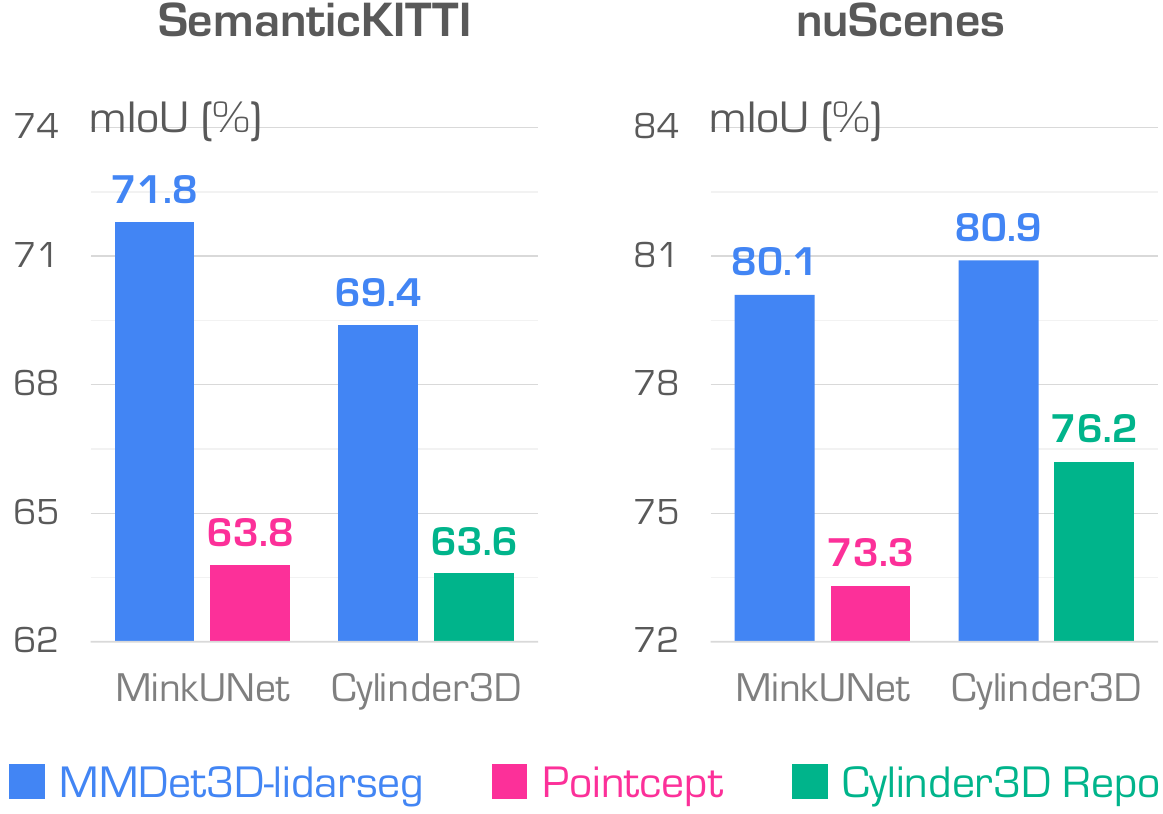}
    \caption{Performance comparisons of state-of-the-art LiDAR segmentation models~\cite{choy2019minkowski,zhu2021cylindrical} from different codebases on the validation sets of the SemanticKITTI~\cite{behley2019semanticKITTI} and nuScenes~\cite{fong2022panoptic-nuScenes} datasets.}
    \label{fig:codebase_compare}
\end{figure}

\begin{table*}[t]
\centering
\caption{An overview of the supported models, sparse convolution backends, and 3D data augmentation techniques from existing LiDAR segmentation codebases. MMDet3D-lidarseg is an abbreviation of our codebase, ``\textcolor{ForestGreen}{\textbf{\checkmark}}'' denotes features that are officially supported.}
\label{tab:model_zoo}
\scalebox{0.92}{
\begin{tabular}{r|p{52pt}<{\centering}|p{43pt}<{\centering}|p{43pt}<{\centering}|p{43pt}<{\centering}|p{43pt}<{\centering}|p{43pt}<{\centering}|p{43pt}<{\centering}}
    \toprule
    & \textbf{\textcolor{lblue}{\includegraphics[height=0.56cm]{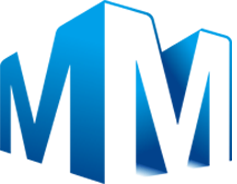} MMDet3D-lidarseg}} & \textbf{\includegraphics[height=0.54cm]{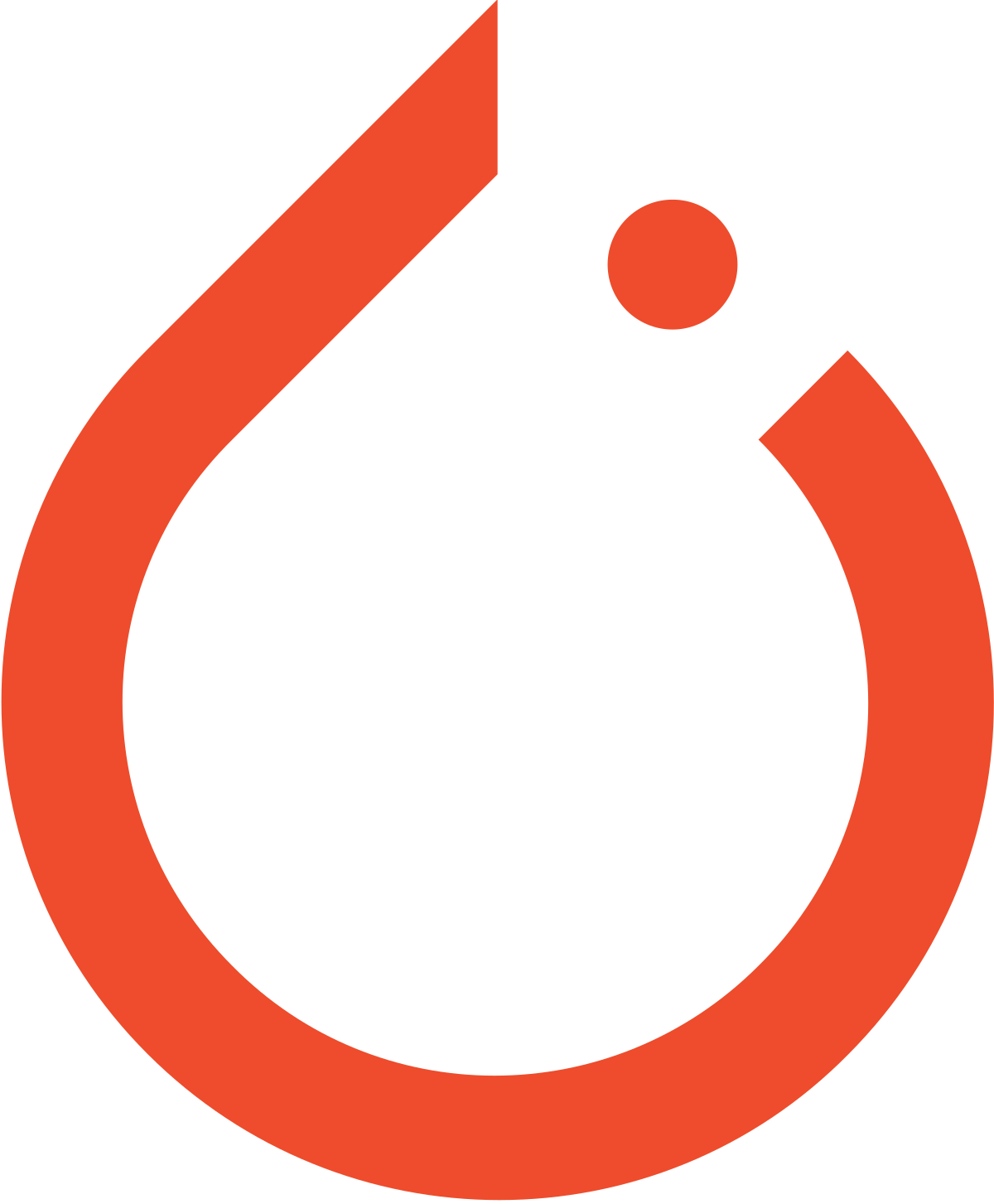} LaserMix} \cite{kong2022laserMix} & \textbf{\includegraphics[height=0.54cm]{figures/pytorch-logo.png} Pointcept} \cite{pointcept2023} & \textbf{\includegraphics[height=0.54cm]{figures/pytorch-logo.png} PCSeg} \cite{openpcseg2023} & \textbf{\includegraphics[height=0.54cm]{figures/pytorch-logo.png} SPVNAS} \cite{tang2020searching} & \textbf{\includegraphics[height=0.54cm]{figures/pytorch-logo.png} Cylin3D} \cite{zhu2021cylindrical} & \textbf{\includegraphics[height=0.54cm]{figures/pytorch-logo.png} PVKD} \cite{pvkd}
    \\\midrule\midrule
    \rowcolor{lblue!9}\multicolumn{8}{l}{\textbf{~~~~~~~~~~~~~~~~~~~~~~~~~~~~~\textcolor{lblue}{Model}}}
    \\
    MinkUNet \cite{choy2019minkowski} & \textcolor{ForestGreen}{\textbf{\textcolor{ForestGreen}{\textbf{\checkmark}}}} & \textcolor{ForestGreen}{\textbf{\checkmark}} & \textcolor{ForestGreen}{\textbf{\checkmark}} & \textcolor{ForestGreen}{\textbf{\checkmark}} & \textcolor{red}{\ding{55}} & \textcolor{red}{\ding{55}} & \textcolor{red}{\ding{55}}
    \\
    % MinkUNet v2 \cite{liu2023uniseg} & \textcolor{ForestGreen}{\textbf{\checkmark}} &  & \textcolor{ForestGreen}{\textbf{\checkmark}} & & \\
    SPVCNN \cite{tang2020searching} & \textcolor{ForestGreen}{\textbf{\checkmark}} & \textcolor{ForestGreen}{\textbf{\checkmark}} & \textcolor{ForestGreen}{\textbf{\checkmark}} & \textcolor{ForestGreen}{\textbf{\checkmark}} & \textcolor{ForestGreen}{\textbf{\checkmark}} & \textcolor{red}{\ding{55}} & \textcolor{red}{\ding{55}}
    \\
    Cylinder3D \cite{zhu2021cylindrical} & \textcolor{ForestGreen}{\textbf{\checkmark}} & \textcolor{ForestGreen}{\textbf{\checkmark}} & \textcolor{red}{\ding{55}} & \textcolor{ForestGreen}{\textbf{\checkmark}} & \textcolor{red}{\ding{55}} & \textcolor{ForestGreen}{\textbf{\checkmark}} & \textcolor{ForestGreen}{\textbf{\checkmark}} 
    \\
    PolarNet \cite{zhou2020polarNet} & \textcolor{ForestGreen}{\textbf{\checkmark}} & \textcolor{ForestGreen}{\textbf{\checkmark}} & \textcolor{red}{\ding{55}} & \textcolor{red}{\ding{55}} & \textcolor{red}{\ding{55}} & \textcolor{red}{\ding{55}} & \textcolor{red}{\ding{55}}
    \\
    CENet \cite{cheng2022cenet} & \textcolor{ForestGreen}{\textbf{\checkmark}} & \textcolor{red}{\ding{55}} & \textcolor{red}{\ding{55}} & \textcolor{red}{\ding{55}} & \textcolor{red}{\ding{55}} & \textcolor{red}{\ding{55}} & \textcolor{ForestGreen}{\textbf{\checkmark}}
    \\
    FRNet \cite{xu2023frnet} & \textcolor{ForestGreen}{\textbf{\checkmark}} & \textcolor{red}{\ding{55}} & \textcolor{red}{\ding{55}} & \textcolor{red}{\ding{55}} & \textcolor{red}{\ding{55}} & \textcolor{red}{\ding{55}} & \textcolor{red}{\ding{55}}
    \\
    \midrule
    \rowcolor{lblue!9}\multicolumn{8}{l}{\textbf{~~~~~~~~~~~~~~~~~~~~~~~~~\textcolor{lblue}{Backend}}}
    \\
    Minkowski Engine \cite{choy2019minkowski} & \textcolor{ForestGreen}{\textbf{\checkmark}} & \textcolor{ForestGreen}{\textbf{\checkmark}} & \textcolor{ForestGreen}{\textbf{\checkmark}} & \textcolor{red}{\ding{55}} & \textcolor{red}{\ding{55}} & \textcolor{red}{\ding{55}} & \textcolor{red}{\ding{55}}
    \\
    SpConv v1 \cite{yan2018second} & \textcolor{ForestGreen}{\textbf{\checkmark}} & \textcolor{ForestGreen}{\textbf{\checkmark}} & \textcolor{red}{\ding{55}} & \textcolor{red}{\ding{55}} & \textcolor{red}{\ding{55}} & \textcolor{ForestGreen}{\textbf{\checkmark}} & \textcolor{ForestGreen}{\textbf{\checkmark}} 
    \\
    SpConv v2 \cite{spconv2022} & \textcolor{ForestGreen}{\textbf{\checkmark}} & \textcolor{red}{\ding{55}} & \textcolor{ForestGreen}{\textbf{\checkmark}} & \textcolor{red}{\ding{55}} & \textcolor{red}{\ding{55}} & \textcolor{red}{\ding{55}} & \textcolor{red}{\ding{55}}
    \\
    TorchSparse \cite{tang2022torchsparse} & \textcolor{ForestGreen}{\textbf{\checkmark}} & \textcolor{ForestGreen}{\textbf{\checkmark}} & \textcolor{ForestGreen}{\textbf{\checkmark}} & \textcolor{ForestGreen}{\textbf{\checkmark}} & \textcolor{ForestGreen}{\textbf{\checkmark}} & \textcolor{red}{\ding{55}} & \textcolor{red}{\ding{55}}
    \\
    TorchSparse++ \cite{tang2023torchsparse++} & \textcolor{ForestGreen}{\textbf{\textcolor{ForestGreen}{\textbf{\checkmark}}}} & \textcolor{red}{\ding{55}} & \textcolor{red}{\ding{55}} & \textcolor{red}{\ding{55}} & \textcolor{ForestGreen}{\textbf{\checkmark}} & \textcolor{red}{\ding{55}} & \textcolor{red}{\ding{55}} 
    \\\midrule
    \rowcolor{lblue!9}\multicolumn{8}{l}{\textbf{~~~~~~~\textcolor{lblue}{Data Augmentation}}}
    \\
    LaserMix \cite{kong2022laserMix} & \textbf{\textcolor{ForestGreen}{\textbf{\checkmark}}} & \textbf{\textcolor{ForestGreen}{\textbf{\checkmark}}} & \textcolor{red}{\ding{55}} & \textcolor{ForestGreen}{\textbf{\checkmark}} & \textcolor{red}{\ding{55}} & \textcolor{red}{\ding{55}} & \textcolor{red}{\ding{55}} 
    \\
    PolarMix \cite{xiao2022polarmix} & \textcolor{ForestGreen}{\textbf{\textcolor{ForestGreen}{\textbf{\checkmark}}}} & \textcolor{red}{\ding{55}} &  \textcolor{red}{\ding{55}} & \textcolor{ForestGreen}{\textbf{\checkmark}} & \textcolor{red}{\ding{55}} & \textcolor{red}{\ding{55}} & \textcolor{red}{\ding{55}}
    \\
    FrustumMix \cite{xu2023frnet} & \textcolor{ForestGreen}{\textbf{\textcolor{ForestGreen}{\textbf{\checkmark}}}} & \textcolor{red}{\ding{55}} & \textcolor{red}{\ding{55}} & \textcolor{red}{\ding{55}} & \textcolor{red}{\ding{55}} & \textcolor{red}{\ding{55}} & \textcolor{red}{\ding{55}} 
    \\
    Test Time Augmentation & \textcolor{ForestGreen}{\textbf{\textcolor{ForestGreen}{\textbf{\checkmark}}}} & \textcolor{red}{\ding{55}} & \textcolor{ForestGreen}{\textbf{\checkmark}} & \textcolor{ForestGreen}{\textbf{\checkmark}} & \textcolor{red}{\textcolor{red}{\ding{55}}} & \textcolor{red}{\ding{55}} & \textcolor{ForestGreen}{\textbf{\checkmark}} 
    \\
    \bottomrule
\end{tabular}
}
\end{table*}

LiDAR (Light Detection and Ranging) is an advanced remote sensing technology that uses laser pulses to measure distances to objects, enabling the capture of high-resolution, three-dimensional information about the environment when mounted on vehicles, drones, or other platforms~\cite{douillard201lidarseg,varney2020dales,geiger2012kitti}. This capability is crucial for applications such as autonomous driving, where understanding complex 3D environments is essential for safe navigation~\cite{behley2021semanticKITTI,liu2024m3net,hong20224dDSNet}. LiDAR provides rich spatial details that complement other sensors like cameras and radars, offering a robust foundation for perceiving and interpreting the surrounding world.

LiDAR segmentation is the process of classifying individual points in a LiDAR-generated point cloud into distinct semantic categories, such as vehicles, pedestrians, and roadways~\cite{behley2019semanticKITTI,fong2022panoptic-nuScenes,caesar2020nuScenes,xiao2022transfer}. This segmentation is fundamental for interpreting the structure and type of objects within the environment, which is critical for situational awareness and decision-making in autonomous systems~\cite{rizzoli2022survey,kong2024lasermix2,li2024place3d,aygun2021pls4d,cheng2023transrvnet,lang2024pattformer}. Effective LiDAR segmentation enhances an autonomous vehicle’s ability to navigate complex scenarios, avoid obstacles, and make informed decisions~\cite{marcuzzi2023mask,zermas2017fast,sirohi2021efficientlps,cao2022monoscene,cao2024pasco,liu2022less}.

Despite its importance, the recent development of LiDAR segmentation models has faced significant challenges. The landscape is fragmented, with numerous standalone and often incompatible codebases~\cite{zhu2021cylindrical,tang2020searching,pvkd}. As shown in \cref{tab:model_zoo}, this fragmentation creates inefficiencies in research and development, making it difficult to perform fair and comprehensive comparisons between different models~\cite{wang2018pointseg,kong2022laserMix,openpcseg2023}. Additionally, many existing solutions often struggle to meet the demands of dynamic and varied autonomous driving environments, which require segmentation frameworks that are flexible, scalable, and robust~\cite{mmdet3d2020,liu2024m3net,kong2024lasermix2,loiseau2022online,chen2021moving,meng2024small,piroli2024label,osep2024sal}.

One major challenge in LiDAR segmentation is the integration of different sparse convolution backends~\cite{tang2022torchsparse,tang2023torchsparse++,spconv2022,yan2018second,choy2019minkowski}, which are essential for efficiently processing the sparse and irregular nature of LiDAR point clouds~\cite{behley2021lps,li2023mseg3d,li2022sdseg3d,li2023tfnet}. Sparse convolutional networks have demonstrated significant improvements in performance and computational efficiency for 3D point cloud processing~\cite{tang2020searching}. However, exploring and comparing these backends within a unified framework has been challenging due to the lack of standardized tools and benchmarks~\cite{mmdet3d2020,openpcseg2023}.

In response to the aforementioned challenges, in this work, we present \textcolor{lblue}{\textbf{MMDetection3D-lidarseg}}, a state-of-the-art toolbox designed to unify, optimize, and streamline the training and benchmarking of LiDAR segmentation models. MMDetection3D-lidarseg integrates multiple advanced features to facilitate cohesive and efficient development, offering a unified framework that enhances the comparability and reproducibility of research findings. Our toolbox supports the exploration and comparison of various sparse convolutional backends, providing a standardized benchmark that improves the performance of widely used segmentation models.

Our design emphasizes several key principles as follows:
\begin{itemize}
    \item \textcolor{lblue}{\textbf{\faUser~Unified Framework:}} By consolidating various LiDAR segmentation models and techniques into a single, comprehensive toolbox, we eliminate the fragmentation of codebases, enabling more efficient development and easier benchmarking of models.

    \item \textcolor{lblue}{\textbf{\faRocket~Optimization \& Efficiency:}} Our toolbox includes optimized implementations of state-of-the-art algorithms, ensuring that models can be trained and evaluated quickly and effectively. This is crucial for both academic research and real-world applications, where time and resource efficiency are paramount.
    
    \item \textcolor{lblue}{\textbf{\faTachometer~Flexibility \& Scalability:}} The MMDetection3D-lidarseg toolkit is designed to be flexible and scalable, capable of adapting to different driving scenarios and handling large-scale point cloud data. This flexibility ensures that the toolbox can meet the evolving needs of researchers and practitioners in the field.

    \item \textcolor{lblue}{\textbf{\faPieChart~Comprehensive Benchmarking:}} To facilitate fair and comprehensive comparisons, we provide a suite of standardized benchmarks that cover a wide range of tasks, including fully-, semi-, and weakly-supervised LiDAR segmentation. These benchmarks help ensure that different models can be evaluated on a common ground, promoting transparency and rigor in research.

    \item \textcolor{lblue}{\textbf{\faCodeFork~Public Availability \& Collaborations:}} By making the codebase and trained models publicly available, we encourage collaboration and open innovation in the field of LiDAR segmentation. This openness helps accelerate the pace of advancements and fosters a community-driven approach to problem-solving.
\end{itemize}

\textcolor{lblue}{\textbf{MMDetection3D-lidarseg}} represents a significant step forward in the development of LiDAR segmentation models, addressing key challenges and setting new standards for research and application. As shown in \cref{fig:codebase_compare}, our implemented models achieve consistent and significant performance improvements over the existing codebases. By providing a comprehensive and flexible platform, our toolbox aims to accelerate innovation in autonomous driving technologies, ultimately contributing to the development of safer and more reliable autonomous systems.
\section{Related Work}
\label{sec:related_work}

\subsection{LiDAR Semantic Segmentation}
Segmenting LiDAR-acquired point clouds into meaningful semantic categories is indispensable for achieving comprehensive 3D scene understanding, especially in the context of autonomous driving~\cite{triess2021survey,gao2021survey,xiao2023survey}. The evolution of LiDAR semantic segmentation methodologies has been characterized by a transition from early hand-crafted features~\cite{weinmann2015semantic,landrieu2017structured} and traditional machine learning techniques~\cite{ester1996dbscan,foschler1981ransac} to the dominance of deep learning approaches~\cite{hu2021sensatUrban,hu2020randla,thomas2019kpconv}. Convolutional neural networks (CNNs), initially developed for image analysis, have been adapted for processing unstructured point cloud data through innovative strategies~\cite{uecker2022analyzing}. One such strategy is voxelization~\cite{choy2019minkowski,tang2020searching,zhu2021cylindrical,hong20224dDSNet}, which transforms point clouds into a structured 3D grid format suitable for standard 3D CNN operations. However, voxelization often incurs high computational costs and memory usage due to the sparsity of point cloud data. Alternatively, projection-based methods~\cite{milioto2019rangenet++,xu2020squeezesegv3,cortinhal2020salsanext,zhou2020polarNet,liong2020amvNet} rasterize point clouds into 2D range images or bird's eye view maps, leveraging the efficiency of 2D CNNs. Despite their efficiency, these methods may sacrifice some 3D spatial information~\cite{jhaldiyal2023survey,ando2023rangevit,kong2023rethinking,xu2023frnet}. The evolution of LiDAR semantic segmentation has been marked by continuous exploration of these methodologies, each offering unique trade-offs in terms of accuracy, computational efficiency, and scalability~\cite{puy23waffleiron,behley2021semanticKITTI,sun2020waymoOpen,unal2022scribbleKITTI}. Recent advancements have focused on integrating multi-modal data sources~\cite{jaritz2020xMUDA,liu2023seal,chen2023towards,liu2023uniseg,yan2022dpass,cheng2021af2S3Net,2023CLIP2Scene,puy2024three,liao2024vlm2scene}, enhancing real-time processing capabilities~\cite{xu2023frnet,zhao2021fidnet,cheng2022cenet,kong2023conDA,chen2021polarStream,triess2020scan}, and improving model robustness under diverse environmental conditions~\cite{kong2023robo3D,kong2023robodepth,hahner2021fog,hahner2022snowfall,xie2023robobev,yu2022benchmarking}. These advancements highlight the dynamic nature of research in LiDAR segmentation, striving to balance the competing demands of precision, efficiency, and robustness~\cite{gasperini2021panoster,cen2022open,yu2022benchmarking}.

\subsection{3D Operators}
Sparse convolution is a fundamental operator in LiDAR segmentation. To understand its significance, we first consider the $N$-dimensional dense convolution. Let $x^{\text{in}}_i$ and $x^{\text{out}}_j$ represent the input and output features, respectively, where $i$ and $j$ are $N$-dimensional coordinates (\eg, $i = (2,1,1)$ in 3D space). The size of the convolution kernel is $K$, with kernel offsets $\Delta^{N}(K)$. The weights of the convolution kernel, $W$, are of shape $K^N \times C^{\text{in}} \times C^{\text{out}}$, where $C^{\text{in}}$ and $C^{\text{out}}$ represent the dimension of input and out features, respectively. For each offset $\delta$ (\eg, $\delta = (0,1,-1) \in \Delta^{3}(3)$), the weights $W$ can be broken down into $K^{N}$ matrices of shape $C^{\text{in}} \times C^{\text{out}}$, denoted as $W_\delta$. Dense convolution with stride $s$ can then be formulated as follows:
\begin{equation}
x^{\text{out}}_i = \sum_{\delta \in \Delta^{N}(K)} x^{\text{in}}_{s \cdot i+\delta} \cdot W_\delta~.
\end{equation}
Here, the bias in convolution is ignored for simplicity. However, in datasets commonly used for LiDAR segmentation, voxelized point cloud samples often have less than $1\%$ of nonzero voxels. This sparsity makes dense convolution inefficient due to the large number of meaningless computations, leading to limited model performance. Sparse convolution addresses this inefficiency by computing only when the input voxel is nonzero. It can be formulated as follows:
\begin{equation}
x^{\text{out}}_i = \sum_{\delta \in \Delta^{N}(K)} \sum_{j} \mathbbm{1}[s\cdot j+\delta ]\cdot x^{\text{in}}_{s \cdot j+\delta }\cdot W_\delta ~,
\end{equation}
where $\mathbbm{1}[\cdot]$ is a binary indicator. The computation of sparse convolution is determined by maps $\mathcal{M}={(i, j, \delta)}$. For given input coordinates and convolution kernel size, sparse convolution first calculates the output coordinates and generates maps $\mathcal{M}$, then performs the convolution operation. In essence, sparse convolution modifies the definition of dense convolution by performing computations only at a sparse set of output locations rather than across the entire feature map. It is arguably the most critical building block for state-of-the-art voxel-based LiDAR segmentation models. Various backends have been developed to efficiently handle sparse convolution tasks, and MMDetection3D-lidarseg supports five prevailing sparse convolution backends: SpConv~\cite{yan2018second}, MinkowskiEngine~\cite{choy2019minkowski}, TorchSparse~\cite{tang2022torchsparse}, SpConv2~\cite{spconv2022}, and TorchSparse++~\cite{tang2023torchsparse++}. We also compare the advantages and disadvantages of these sparse convolutions in the experimental section, aiming to assist researchers in selecting the most suitable library for their specific needs.

\subsection{3D Data Augmentations}
Data augmentation plays a crucial role in enhancing the performance and generalizability of deep learning models, especially in domains characterized by limited training data or substantial variance in operational conditions~\cite{xiao2022polarmix, kong2022laserMix, xu2023frnet}. In the context of LiDAR segmentation, 3D data augmentations are essential for simulating a wide range of scenarios and environmental conditions that autonomous vehicles may encounter~\cite{kong2023robo3D}. Techniques such as random rotations, scaling, and jittering of point clouds help models become robust to variations in object orientations and sizes. More sophisticated methods, including synthetic occlusion, simulating varying sensor ranges, and generating adversarial examples, further enrich the model's exposure to challenging conditions, thereby improving its real-world performance~\cite{kong2024calib3d,li2024place3d}. Additionally, domain adaptation techniques aim to bridge the gap between synthetic and real-world data, enabling models trained primarily on simulated environments to transfer their capabilities to real-world settings~\cite{kong2023conDA,jaritz2020xMUDA,li2022coarse3D,xu2024visual,michele2024saluda,sautier2024bevcontrast,boulch2023also}. Continuous innovation in 3D data augmentation techniques is pivotal for advancing the state-of-the-art in LiDAR segmentation, ensuring models are not only accurate but also resilient to the diverse challenges they face in real-world applications~\cite{kong2023robo3D,xie2023robobev}. These augmentations, coupled with robust training frameworks like MMDetection3D-lidarseg, provide a comprehensive toolkit for developing next-generation LiDAR segmentation models capable of meeting the demands of modern autonomous systems~\cite{kong2024lasermix2}.
\section{MMDetection3D-lidarseg}
\label{sec:supports}

MMDetection3D-lidarseg includes high-quality implementations of popular LiDAR semantic segmentation models and sparse convolution backends. A summary of supported models and backends, compared to other codebases, is provided in \cref{tab:model_zoo}. MMDetection3D-lidarseg supports a wider range of models and backends than other codebases, offering unparalleled flexibility and comprehensiveness. The following is a brief introduction to the datasets, models, data augmentation techniques, and sparse convolution backends that we support.

\subsection{Datasets}
\begin{itemize}
    \item \textbf{SemanticKITTI}~\cite{behley2019semanticKITTI} is a pioneering dataset for LiDAR segmentation, offering an extensive collection of annotated 3D point clouds derived from real-world driving scenarios. It extends the KITTI Vision Benchmark Suite~\cite{geiger2012kitti} by providing dense, per-point annotations across a wide range of categories, including various types of vehicles, pedestrians, and both natural and urban structures. Notably, this dataset features temporal consistency, capturing the dynamic nature of urban environments through sequences of scans over time. The extensive coverage and challenging scenarios make SemanticKITTI a fundamental resource for advancing and benchmarking LiDAR segmentation models in the context of autonomous driving.

    \item \textbf{nuScenes-lidarseg}~\cite{fong2022panoptic-nuScenes}, a variant of the well-known nuScenes \cite{caesar2020nuScenes} dataset, serves as a diverse and large-scale benchmark for semantic segmentation in autonomous vehicle sensor data. This dataset includes detailed annotations for a variety of object classes across numerous scenes captured in different urban environments and weather conditions, utilizing LiDAR, radar, and camera data. The complexity and diversity of nuScenes, featuring night scenes and adverse weather conditions, present significant challenges for segmentation models, making it an invaluable resource for testing robustness and versatility. Additionally, its multi-modal nature supports research into sensor fusion techniques, enhancing its contribution to the development of comprehensive perception systems for autonomous driving.

    \item \textbf{ScribbleKITTI}~\cite{unal2022scribbleKITTI} introduces an innovative approach to dataset annotation for LiDAR segmentation. It shares the same scenes with SemanticKITTI~\cite{behley2019semanticKITTI} but is annotated with line scribbles rather than dense, per-point labeling. This method significantly reduces the annotation effort while still providing high-quality training data for weakly-supervised learning algorithms. This efficiency of scribble-based annotations enables the creation of extensive datasets with less manual effort, facilitating the development and testing of advanced segmentation models in autonomous driving applications.
\end{itemize}

\subsection{Models}
\begin{itemize}
    \item \textbf{MinkUNet}~\cite{choy2019minkowski} is a highly efficient model for LiDAR segmentation, built on the Minkowski Engine. It employs sparse convolutional operations to efficiently process point cloud data, extending the U-Net architecture into 3D spaces. This design enables the segmentation of large-scale point clouds by leveraging the inherent sparsity of LiDAR data, significantly enhancing computational efficiency.
    
    \item \textbf{SPVCNN}~\cite{tang2020searching} combines the strengths of point-based and voxel-based processing methods to achieve high efficiency and accuracy. This hybrid approach provides a scalable solution that adapts to the density variations of point cloud data, ensuring detailed feature extraction while maintaining computational efficiency across diverse scenes.

    \item \textbf{Cylinder3D}~\cite{zhu2021cylindrical} introduces a cylindrical partitioning strategy that aligns more naturally with the distribution of point clouds captured by circular LiDAR scans. This approach enhances segmentation performance, particularly for cylindrical objects like poles and trees, by preserving their geometric integrity.

    \item \textbf{CENet}~\cite{cheng2022cenet} focuses on capturing and utilizing contextual information from the range view of LiDAR point clouds for efficient segmentation, with multi-level feature fusion and . 

    \item \textbf{PolarNet}~\cite{zhou2020polarNet} transforms LiDAR point clouds into a polar bird's eye view representation, facilitating efficient processing through partitioning the space into polar bins. This representation is particularly advantageous for LiDAR data, aligning with the sensor's native data structure and simplifying the detection of objects in 360-degree surroundings.

    \item \textbf{FRNet}~\cite{xu2023frnet} introduces a novel Frustum-Range representation for scalable LiDAR segmentation and proposes two novel 3D data augmentation techniques, FrustumMix and RangeInterpolation, to assist in training robust and generalizable models. It achieves competitive performance while still maintaining promising scalability for real-time LiDAR segmentation.
\end{itemize}

Although the architectures of various LiDAR segmentation models differ, they can generally be categorized into voxel-based and projection-based segmentors, each sharing common components~\cite{kong2023rethinking,ando2023rangevit,yan2022dpass}. Voxel-based segmentors~\cite{choy2019minkowski,zhu2021cylindrical,tang2020searching} typically consist of a voxelization module, a voxel feature encoder, a 3D backbone, and a decode head. These components work together to convert raw point cloud data into a structured voxel grid, extract meaningful features, process these features through a deep 3D neural network, and finally decode the processed features into semantic labels. Conversely, projection-based segmentors~\cite{cheng2022cenet,xu2023frnet} include a projection module, a pixel feature encoder, a 2D backbone, and a decode head. The projection module converts the 3D point cloud into a 2D image representation, such as a range image or bird's eye view map~\cite{milioto2019rangenet++,zhou2020polarNet}. The pixel feature encoder and 2D backbone then process this image to extract features, which are subsequently decoded into semantic labels~\cite{wang2018pointseg,xu2020squeezesegv3}.

\begin{figure}[t]
    \centering
    \includegraphics[width=\linewidth,scale=1.00]{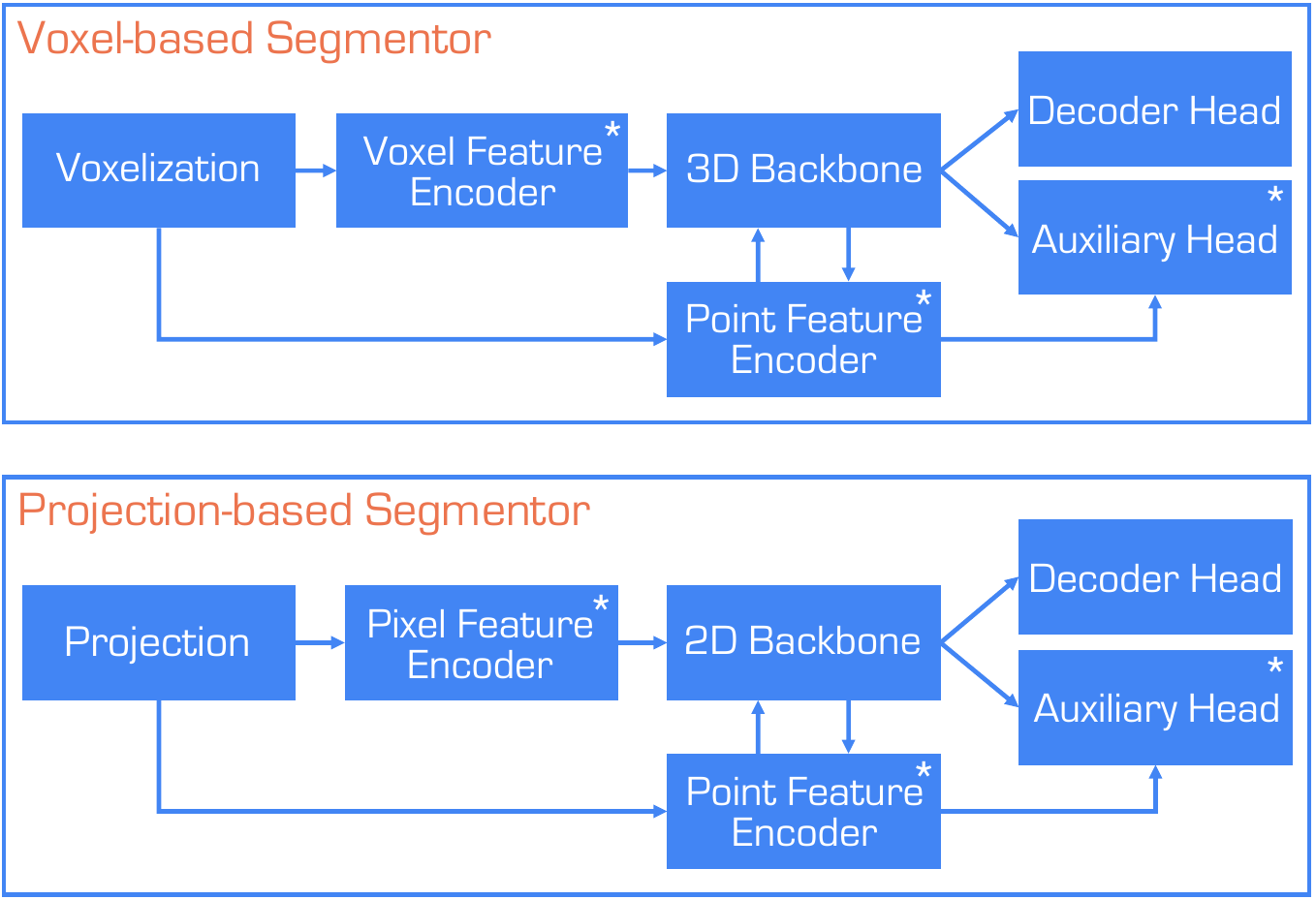}
    \caption{Overview of voxel-based and projection-based LiDAR segmentors illustrated with abstractions in the MMDetection3D-lidarseg codebase. Modules marked with $*$ are optional.}
    \label{fig:framework}
\end{figure}

Both types of segmentors may also incorporate a point feature encoder to directly process raw point cloud data and an auxiliary head to provide additional outputs or intermediate supervision, enhancing the model's overall performance and robustness~\cite{xu2023frnet,zhu2021cylindrical,kong2023robo3D}. With these abstractions, the framework of voxel-based and projection-based segmentors is illustrated in \cref{fig:framework}. The modular design of these segmentors allows researchers to develop their own models by creating new components and assembling them with existing ones. This flexibility not only accelerates the development process but also facilitates experimentation and innovation in the field of LiDAR segmentation.

\subsection{Data Augmentation Techniques}
\begin{itemize}
    \item \textbf{LaserMix}~\cite{kong2022laserMix} introduces a novel augmentation technique specifically designed for LiDAR point clouds, inspired by the concept of mixup for image data. This technique involves mixing multiple point clouds along the azimuth and inclination directions, significantly enhancing the model's ability to learn generalized features and improve robustness against overfitting.
    
    \item \textbf{PolarMix}~\cite{xiao2022polarmix} generates synthetic samples by mixing polar representations of LiDAR point clouds. This method is particularly effective for augmenting data in scenarios where the spatial relationships and orientations of objects are critical for understanding the scene. 

    \item \textbf{FrustumMix}~\cite{xu2023frnet} enhances context awareness by swapping frustum regions from two scenes. This method splits two point clouds into several non-overlapping frustum regions along the inclination or azimuth direction. FrustumMix maintains the invariant structure within the frustum while enhancing the contextual relationships between frustums.

    \item \textbf{Test Time Augmentation (TTA)}, typically applied during model inference, enhances segmentation accuracy by applying various transformations (\eg, rotations, scaling) to the input data at test time and aggregating the predictions. TTA mitigates the impact of noise and variations in the point cloud data, resulting in more consistent and reliable segmentation outcomes, particularly in challenging or ambiguous scenes. However, it requires significantly more time during inference, making it less suitable for real-time applications such as autonomous driving perception.
\end{itemize}

\subsection{Sparse Convolution Backends}
\begin{itemize}
    \item \textbf{SpConv}~\cite{yan2018second} can be regarded as a vanilla implementation of the sparse convolution. SpConv employs the gather-GEMM-scatter dataflow on GPUs, which is weight-stationary and features an outer host loop over kernel offsets. For each offset, it computes a kernel map, gathers all input features into a $K^2 \times C^{\text{in}}$ matrix in DRAM, and multiplies the weights using dense matrix multiplication, handled by existing vendor libraries like cuBLAS and cuDNN. SpConv then scatters the results back to output positions. This implementation only needs to optimize scatter and gather in CUDA. However, this data flow is fundamentally inefficient due to the lack of overlap between computation and memory access.
    
    \item \textbf{MinkowskiEngine}~\cite{choy2019minkowski} takes a fetch-on-demand approach and merges the gather, matrix multiplication, and scatter kernel calls into a single CUDA kernel. Instead of materializing the $K^2 \times C^{\text{in}}$ matrix gather buffer in DRAM, it fetches input features on demand into the L1 shared memory, performs matrix multiplication in the on-chip storage, and directly scatters the partial sums (residing in the register file) to corresponding outputs position without first instantiating them in a DRAM scatter buffer. This method overlaps computation with memory access and saves DRAM writes to gather and scatter buffers. However, it cannot avoid DRAM writes to the final output tensor, resulting in $4\times$ to $10\times$ larger write-back traffic than the theoretical optimal value. Furthermore, the block-fused fetch-on-demand dataflow suffers from write-back contentions between different threads. 
    
    \item \textbf{TorchSparse}~\cite{tang2022torchsparse} directly optimizes the two bottlenecks of the vanilla gather-GEMM-scatter dataflow: irregular computation and data movement. Due to sparse convolutions with odd kernel size and a stride of 1, the maps corresponding to kernel offsets $(a, b, c)$ always have the same size as the maps corresponding to the symmetric kernel offset $(-a, -b, -c)$, TorchSparse applies adaptive matrix multiplication grouping to trade computation for better regularity, achieving a $1.4\times$ to $1.5\times$ speedup for matrix multiplication. TorchSparse also optimizes data movement by adopting vectorized, quantized, and fused locality-aware memory access, reducing the memory movement cost by $2.7\times$.

    \item \textbf{SpConv v2}~\cite{spconv2022} extends the well-known implicit GEMM formulation for 2D convolution to 3D sparse convolution. The sparse convolution workload is equivalent to a dense GEMM. Similar to fetch-on-demand, implicit GEMM overlaps computation with memory access, hiding memory latency through pipelining. Like ``\texttt{im2col}'' in 2D convolution, an implicit GEMM implementation is output-stationary, achieving the theoretical minimum DRAM write-back traffic. However, despite lower DRAM traffic compared to fetch-on-demand, implicit GEMM has non-negligible redundant computation. To address this issue, SpConv v2 excludes unsorted implicit GEMM in their design space and uses bitmask sorting to minimize computation overhead. Each output point is assigned a $K^2$ dimensional bitmask indicating the presence of its neighbors. These bitmasks are treated as numbers and sorted, adjusting the computation order for different outputs accordingly. In practical applications, sorting can reduce redundant computation by up to $3\times$.

    \item \textbf{TorchSparse++}~\cite{tang2023torchsparse++} simultaneously supports gather-GEMM-scatter, fetch-on-demand, and sorted implicit GEMM dataflows. Overlapped computation and memory access can be achieved with relatively low engineering complexity compared to SpConv2. TorchSparse++ creates a highly efficient Sparse Kernel Generator that generates performant sparse convolution kernels at less than one-tenth of the engineering cost of current state-of-the-art systems. Additionally, TorchSparse++ designed the Sparse Autotuner, which extends the design space of existing sparse convolution libraries and searches for the best dataflow configurations for training and inference workloads.
\end{itemize}

\section{Experiments}
\label{sec:benchmark}

In this section, we embark on comprehensive benchmarks of diverse LiDAR segmentation models facilitated by MMDetection3D-lidarseg. Firstly, we provide an overview of the datasets, metrics, and detailed implementations employed within the codebase. Subsequently, we delve into experimental results from distinct setups, encompassing fully-, semi-, and weakly-supervised learning paradigms. Finally, we delve into a series of ablation studies, aiming to scrutinize and compare the various components facilitated within the codebase. Through these rigorous evaluations, we aim to provide valuable insights into the performance and efficacy of LiDAR segmentation modes, contributing to advancements in this critical field.

\begin{table}[t]
    \centering
    \caption{Results of state-of-the-art LiDAR segmentation models~\cite{choy2019minkowski,zhu2021cylindrical,zhou2020polarNet,cheng2022cenet,xu2023frnet,tang2020searching} on the validation sets of SemanticKITTI~\cite{behley2019semanticKITTI} and nuScenes~\cite{fong2022panoptic-nuScenes}. \textbf{Mix} denote a combination of LaserMix~\cite{kong2022laserMix} and PolarMix~\cite{xiao2022polarmix} or FrustumMix \cite{xu2023frnet}. \textbf{TTA} denotes test time augmentation. All scores are given in percentage (\%). The \textbf{best} scores under each metric are highlighted in \textbf{bold}.}
    \scalebox{0.92}{
    \begin{tabular}{c|p{17pt}<{\centering}p{17pt}<{\centering}|p{22pt}<{\centering}p{22pt}<{\centering}|p{22pt}<{\centering}p{22pt}<{\centering}}
        \toprule
        \multirow{2}{*}{\textbf{Model}} & \multirow{2}{*}{\textbf{Mix}} & \multirow{2}{*}{\textbf{TTA}}  &  \multicolumn{2}{c|}{\textbf{SemKITTI}} & \multicolumn{2}{c}{\textbf{nuScenes}}
        \\
        & & & mIoU & mAcc & mIoU & mAcc
        \\\midrule\midrule
        \rowcolor{lblue!9}\multicolumn{7}{l}{\textcolor{lblue}{\textbf{Representation: Voxel}}}
        \\\midrule
        \multirow{3}{*}{MinkUNet} & \textcolor{red}{\ding{55}} & \textcolor{red}{\ding{55}} & $66.9$ & $92.4$ & $76.4$ & $94.1$
        \\
        & \textcolor{ForestGreen}{\textbf{\checkmark}} & \textcolor{red}{\ding{55}} & $70.4$ & $92.8$ & $77.6$ & $94.2$ 
        \\
        & \textcolor{ForestGreen}{\textbf{\checkmark}} & \textcolor{ForestGreen}{\textbf{\checkmark}} & $\mathbf{71.8}$ & $\mathbf{93.2}$ & $80.1$ & $94.7$ 
        \\\midrule
        \multirow{3}{*}{Cylinder3D} & \textcolor{red}{\ding{55}} & \textcolor{red}{\ding{55}} & $63.7$ & $91.0$ & $75.8$ & $93.7$
        \\
        & \textcolor{ForestGreen}{\textbf{\checkmark}} & \textcolor{red}{\ding{55}} & $67.0$ & $91.7$ & $79.3$ & $94.3$
        \\
        & \textcolor{ForestGreen}{\textbf{\checkmark}} & \textcolor{ForestGreen}{\textbf{\checkmark}} & $69.4$ & $92.5$ & $\mathbf{80.9}$ & $\mathbf{94.5}$
        \\\midrule
        \rowcolor{lblue!9}\multicolumn{7}{l}{\textcolor{lblue}{\textbf{Representation: Bird's Eye View}}}
        \\\midrule
        \multirow{3}{*}{PolarNet} & \textcolor{red}{\ding{55}} & \textcolor{red}{\ding{55}} & $57.2$ & $91.0$ & $71.7$ & $93.1$ 
        \\
        & \textcolor{ForestGreen}{\textbf{\checkmark}} & \textcolor{red}{\ding{55}} & $60.1$ & $91.3$ & $72.3$ & $93.2$ 
        \\
        & \textcolor{ForestGreen}{\textbf{\checkmark}} & \textcolor{ForestGreen}{\textbf{\checkmark}} & $60.7$ & $91.7$ & $72.3$ & $93.4$ 
        \\\midrule
        \rowcolor{lblue!9}\multicolumn{7}{l}{\textcolor{lblue}{\textbf{Representation: Range View}}}
        \\\midrule
        \multirow{2}{*}{CENet} & \textcolor{red}{\ding{55}} & \textcolor{red}{\ding{55}} & $61.9$ & $90.3$ & - & -
        \\
        & \textcolor{ForestGreen}{\textbf{\checkmark}} & \textcolor{red}{\ding{55}} & $62.2$ & $90.5$ & - & - 
        \\\midrule
        \multirow{3}{*}{FRNet} & \textcolor{red}{\ding{55}} & \textcolor{red}{\ding{55}} & $64.1$ & $92.2$ & $76.8$ & $93.4$ 
        \\
        & \textcolor{ForestGreen}{\textbf{\checkmark}} & \textcolor{red}{\ding{55}} & $67.6$ & $92.3$ & $77.7$ & $94.0$ 
        \\
        & \textcolor{ForestGreen}{\textbf{\checkmark}} & \textcolor{ForestGreen}{\textbf{\checkmark}} & $68.7$ & $92.5$ & $79.0$ & $94.3$ 
        \\\midrule
        \rowcolor{lblue!9}\multicolumn{7}{l}{\textcolor{lblue}{\textbf{Representation: Fusion}}}
        \\\midrule
        \multirow{2}{*}{SPVCNN} & \textcolor{red}{\ding{55}} & \textcolor{red}{\ding{55}} & $66.4$ & $92.5$ & $76.0$ & $93.9$ 
        \\
        & \textcolor{ForestGreen}{\textbf{\checkmark}} & \textcolor{red}{\ding{55}} & $68.4$ & $92.3$ & $77.0$ & $94.1$ 
        \\
        \bottomrule
    \end{tabular}}
    \label{tab:main_results}
\end{table}
\begin{table*}[t]
\caption{Results of data-efficient learning algorithms~\cite{MeanTeacher,CBST,CutMix-Seg,CPS,kong2022laserMix,xu2023frnet,unal2022scribbleKITTI,li23lim3d} with state-of-the-art LiDAR segmentation backbones~\cite{zhao2021fidnet,xu2023frnet,zhou2020polarNet,zhu2021cylindrical,choy2019minkowski,tang2020searching} on the semi-supervised learning benchmarks of the SemanticKITTI~\cite{behley2019semanticKITTI}, nuScenes~\cite{fong2022panoptic-nuScenes}, and ScribbleKITTI~\cite{unal2022scribbleKITTI} datasets, respectively. All methods are tested on the official validation sets. The labeled data are uniformly sampled with the 1\%, 10\%, 20\%, and 50\% quota from the training set of the original dataset. \textit{Sup.-Only} denotes training with the labeled data only, while the data-efficient learning methods use both labeled and unlabeled data. $^\dagger$Lim3D~\cite{li23lim3d} adopts a different data sampling strategy instead of uniform sampling. All scores are given in percentage (\%). The \textbf{best} scores under each metric are highlighted in \textbf{bold}.}
\setlength{\tabcolsep}{4pt}
\centering\scalebox{0.92}{
\begin{tabular}{r|r|c|p{21pt}<{\centering}p{21pt}<{\centering}p{21pt}<{\centering}p{21pt}<{\centering}|p{21pt}<{\centering}p{21pt}<{\centering}p{21pt}<{\centering}p{21pt}<{\centering}|p{21pt}<{\centering}p{21pt}<{\centering}p{21pt}<{\centering}p{21pt}<{\centering}}
\toprule
\multirow{2}{*}{\textbf{Model}} & \multirow{2}{*}{\textbf{Venue}} & \multirow{2}{*}{\textbf{Backbone}} & \multicolumn{4}{c|}{\textbf{SemanticKITTI}} &
\multicolumn{4}{c|}{\textbf{nuScenes}} & \multicolumn{4}{c}{\textbf{ScribbleKITTI}}
\\
& & & 1\% & 10\% & 20\% & 50\% & 1\% & 10\% & 20\% & 50\% & 1\% & 10\% & 20\% & 50\%
\\\midrule\midrule
% \multirow{9.5}{*}{\rotatebox{90}{Range View}} &
\rowcolor{lblue!9}\multicolumn{15}{l}{\textcolor{lblue}{\textbf{Representation: Range View}}}
\\\midrule
% \textit{Sup.-Only} & - & \multirow{6}{*}{FIDNet} & \cellcolor{gray!10}$36.2$ & \cellcolor{gray!10}$52.2$ & \cellcolor{gray!10}$55.9$ & \cellcolor{gray!10}$57.2$ & \cellcolor{gray!10}$38.3$ & \cellcolor{gray!10}$57.5$ & \cellcolor{gray!10}$62.7$ & \cellcolor{gray!10}$67.6$ &  \cellcolor{gray!10}$33.1$ & \cellcolor{gray!10}$47.7$ & \cellcolor{gray!10}$49.9$ & \cellcolor{gray!10}$52.5$
% \\
% MeanTeacher~\cite{MeanTeacher} & \small{NeurIPS'17} & & $37.5$ & $53.1$ & $56.1$ & $57.4$ & $42.1$ & $60.4$ & $65.4$ & $69.4$ & $34.2$ & $49.8$ & $51.6$ & $53.3$
% \\
% CBST~\cite{CBST} & \small{ECCV'18} & & $39.9$ & $53.4$ & $56.1$ & $56.9$ & $40.9$ & $60.5$ & $64.3$ & $69.3$ & $35.7$ & $50.7$ & $52.7$ & $54.6$
% \\
% CutMix-Seg~\cite{CutMix-Seg} & \small{BMVC'20} & & $37.4$ & $54.3$ & $56.6$ & $57.6$ & $43.8$ & $63.9$ & $64.8$ & $69.8$ & $36.7$ & $50.7$ & $52.9$ & $54.3$
% \\
% CPS~\cite{CPS} & \small{CVPR'21} & & $36.5$ & $52.3$ & $56.3$ & $57.4$ & $40.7$ & $60.8$ & $64.9$ & $68.0$ & $33.7$ & $50.0$ & $52.8$ & $54.6$
% \\
% LaserMix~\cite{kong2022laserMix} & \small{CVPR'23} & & $43.4$ & $58.8$ & $59.4$ & $61.4$ & $49.5$ & $68.2$ & $70.6$ & $73.0$ & $38.3$ & $54.4$ & $55.6$ & $58.7$
% \\\midrule
\textit{Sup.-Only} & - & \multirow{3}{*}{FRNet} & \cellcolor{gray!10}$44.9$ & \cellcolor{gray!10}$60.4$ & \cellcolor{gray!10}$61.8$ & \cellcolor{gray!10}$63.1$ & \cellcolor{gray!10}$51.9$ & \cellcolor{gray!10}$68.1$ & \cellcolor{gray!10}$70.9$ & \cellcolor{gray!10}$74.6$ & \cellcolor{gray!10}$42.4$ & \cellcolor{gray!10}$53.5$ & \cellcolor{gray!10}$55.1$ & \cellcolor{gray!10}$57.0$
\\
LaserMix~\cite{kong2022laserMix} & \small{CVPR'23} & & $52.9$ & $62.9$ & $63.2$ & $65.0$ & $58.7$ & $71.5$ & $72.3$ & $75.0$ & $45.8$ & $56.8$ & $57.7$ & $59.0$
\\
FrustumMix~\cite{xu2023frnet} & \small{arXiv'23} & & $55.8$ & $64.8$ & $65.2$ & $65.4$ & $61.2$ & $72.2$ & $74.6$ & $75.4$ & $46.6$ & $57.0$ & $59.5$ & $61.2$
\\\midrule\midrule
% \multirow{3}{*}{\rotatebox{90}{BEV}} & 
\rowcolor{lblue!9}\multicolumn{15}{l}{\textcolor{lblue}{\textbf{Representation: Bird's Eye View}}}
\\\midrule
\textit{Sup.-Only} & - & \multirow{3}{*}{PolarNet} & \cellcolor{gray!10}$45.1$ & \cellcolor{gray!10}$54.6$ & \cellcolor{gray!10}$55.6$ & \cellcolor{gray!10}$56.5$ & \cellcolor{gray!10}$50.9$ & \cellcolor{gray!10}$67.5$ & \cellcolor{gray!10}$69.5$ & \cellcolor{gray!10}$71.0$ & \cellcolor{gray!10}$42.6$ & \cellcolor{gray!10}$52.8$ & \cellcolor{gray!10}$53.4$ & \cellcolor{gray!10}$54.4$
\\
MeanTeacher~\cite{MeanTeacher} & \small{NeurIPS'17} & & $47.4$ & $55.6$ & $56.6$ & $57.1$ & $51.9$ & $68.1$ & $69.7$ & $71.1$ & $43.7$ & $53.4$ & $54.$4 & $54.9$
\\
LaserMix~\cite{kong2022laserMix} & \small{CVPR'23} & & $51.0$ & $57.7$ & $58.6$ & $60.0$ & $54.0$ & $69.5$ & $70.8$ & $71.9$ & $45.7$ & $55.5$ & $56.0$ & $56.6$
\\\midrule\midrule
% \multirow{12}{*}{\rotatebox{90}{Voxel}} &
\rowcolor{lblue!9}\multicolumn{15}{l}{\textcolor{lblue}{\textbf{Representation: Voxel}}}
\\\midrule
\textit{Sup.-Only} & - & \multirow{5}{*}{Cylinder3D} & \cellcolor{gray!10}$45.4$ & \cellcolor{gray!10}$56.1$ & \cellcolor{gray!10}$57.8$ & \cellcolor{gray!10}$58.7$ & \cellcolor{gray!10}$50.9$ & \cellcolor{gray!10}$65.9$ & \cellcolor{gray!10}$66.6$ & \cellcolor{gray!10}$71.2$ &  \cellcolor{gray!10}$39.2$ & \cellcolor{gray!10}$48.0$ & \cellcolor{gray!10}$52.1$ & \cellcolor{gray!10}$53.8$
\\
MeanTeacher~\cite{MeanTeacher} & \small{NeurIPS'17} & & $45.4$ & $57.1$ & $59.2$ & $60.0$ & $51.6$ & $66.0$ & $67.1$ & $71.7$ & $41.0$ & $50.1$ & $52.8$ & $53.9$
\\
% CBST~\cite{CBST} & \small{ECCV'18} & & $48.8$ & $58.3$ & $59.4$ & $59.7$ & $53.0$ & $66.5$ & $69.6$ & $71.6$ & $41.5$ & $50.6$ & $53.3$ & $54.5$
% \\
% CPS~\cite{CPS} & \small{CVPR'21} & & $46.7$ & $58.7$ & $59.6$ & $60.5$ & $52.9$ & $66.3$ & $70.0$ & $72.5$ &  $41.4$ & $51.8$ & $53.9$ & $54.8$
% \\
% CRB~\cite{unal2022scribbleKITTI} & \small{CVPR'22} & & - & $58.7$ & $59.1$ & $60.9$ & - & - & - & - & - & $54.2$ & $56.5$ & $58.9$
% \\
LaserMix~\cite{kong2022laserMix} & \small{CVPR'23} & & $50.6$ & $60.0$ & $61.9$ & $62.3$ & $55.3$ & $69.9$ & $71.8$ & $73.2$ & $44.2$ & $53.7$ & $55.1$ & $56.8$
\\
LiM3D$^\dagger$~\cite{li23lim3d} & \small{CVPR'23} & & - & $61.6$ & $62.6$ & $62.8$ & - & - & - & - & - & $60.3$ & $60.5$ & $60.9$
\\
% & ImageTo360~\cite{} & \small{ICCVW'23} & & $54.5$ & $58.6$ & $61.4$ & $64.2$ & - & - & - & - & - & - & - & -
% \\
FrustumMix~\cite{xu2023frnet} & \small{arXiv'23} & & $55.7$ & $62.5$ & $63.0$ & $64.9$ & $60.0$ & $70.0$ & $72.6$ & $74.1$ & $45.6$ & $55.7$ & $58.2$ & $60.8$
\\\midrule
\textit{Sup.-Only} & - & \multirow{3}{*}{MinkUNet} & \cellcolor{gray!10}$53.9$ & \cellcolor{gray!10}$64.0$ & \cellcolor{gray!10}$64.6$ & \cellcolor{gray!10}$65.4$ & \cellcolor{gray!10}$58.3$ & \cellcolor{gray!10}$71.0$ & \cellcolor{gray!10}$73.0$ & \cellcolor{gray!10}$75.1$ & \cellcolor{gray!10}$48.6$ & \cellcolor{gray!10}$57.7$ & \cellcolor{gray!10}$58.5$ & \cellcolor{gray!10}$60.0$
\\
MeanTeacher~\cite{MeanTeacher} & \small{NeurIPS'17} & & $56.1$ & $64.7$ & $65.4$ & $66.0$ & $60.1$ & $71.7$ & $73.4$ & $75.2$ & $49.7$ & $59.4$ & $60.0$ & $61.7$
\\
LaserMix~\cite{kong2022laserMix} & \small{CVPR'23} & & $\mathbf{60.9}$ & $\mathbf{66.6}$ & $\mathbf{67.2}$ & $\mathbf{68.0}$ & $62.8$ & $73.6$ & $\mathbf{74.8}$ & $\mathbf{76.1}$ & $\mathbf{57.2}$ & $\mathbf{61.1}$ & $\mathbf{61.4}$ & $\mathbf{62.4}$
\\\midrule\midrule
% \multirow{3}{*}{\rotatebox{90}{Fusion}} & 
\rowcolor{lblue!9}\multicolumn{15}{l}{\textcolor{lblue}{\textbf{Representation: Fusion}}}
\\\midrule
\textit{Sup.-Only} & - & \multirow{3}{*}{SPVCNN} & \cellcolor{gray!10}$52.7$ & \cellcolor{gray!10}$64.1$ & \cellcolor{gray!10}$64.5$ & \cellcolor{gray!10}$65.1$ & \cellcolor{gray!10}$57.9$ & \cellcolor{gray!10}$71.7$ & \cellcolor{gray!10}$73.0$ & \cellcolor{gray!10}$74.6$ & \cellcolor{gray!10}$47.2$ & \cellcolor{gray!10}$57.3$ & \cellcolor{gray!10}$58.2$ & \cellcolor{gray!10}$58.8$
\\
MeanTeacher~\cite{MeanTeacher} & \small{NeurIPS'17} & & $54.4$ & $64.8$ & $65.2$ & $65.7$ & $59.4$ & $72.5$ & $73.1$ & $74.7$ & $49.9$ & $58.3$ & $58.6$ & $59.1$
\\
LaserMix~\cite{kong2022laserMix} & \small{CVPR'23} & & $60.3$ & $\mathbf{66.6}$ & $67.0$ & $67.6$ & $\mathbf{63.2}$ & $\mathbf{74.1}$ & $74.6$ & $75.8$ & $57.1$ & $60.8$ & $60.7$ & $61.0$
\\\bottomrule
\end{tabular}}
\label{tab:benchmark_semi}
\end{table*}

\subsection{Experimental Settings}
\noindent\textbf{Datasets.}
We conduct comprehensive experiments on three popular LiDAR-based semantic segmentation benchmarks. \textit{\textbf{SemanticKITTI}}~\cite{behley2019semanticKITTI} is a large-scale outdoor dataset consisting of $22$ sequences collected from various scenarios in Karlsruhe, Germany. Typically, sequences $00$ to $10$ are used for training, while sequence $08$ is used for validation. All points are annotated with $28$ classes, and a merged set of $19$ classes is used for evaluation. \textit{\textbf{nuScenes}}~\cite{fong2022panoptic-nuScenes} is a multi-modal dataset widely used in autonomous driving. It is collected around streets in Boston and Singapore and features sparser LiDAR points. The dataset includes $1,000$ driving scenes, with $750$ scenes for training, $150$ scenes for validation, and $100$ scenes for testing. It is annotated with $32$ semantic categories, and a merged set of $16$ labels is used for evaluation. \textit{\textbf{ScribbleKITTI}}~\cite{unal2022scribbleKITTI} shares the same scenes as SemanticKITTI~\cite{behley2019semanticKITTI} but is annotated with line scribbles. It contains $189$ million labeled points, with approximately $8.06\%$ labeled points for training.

\begin{figure}[t]
    \centering
    \includegraphics[width=\linewidth,scale=1.00]{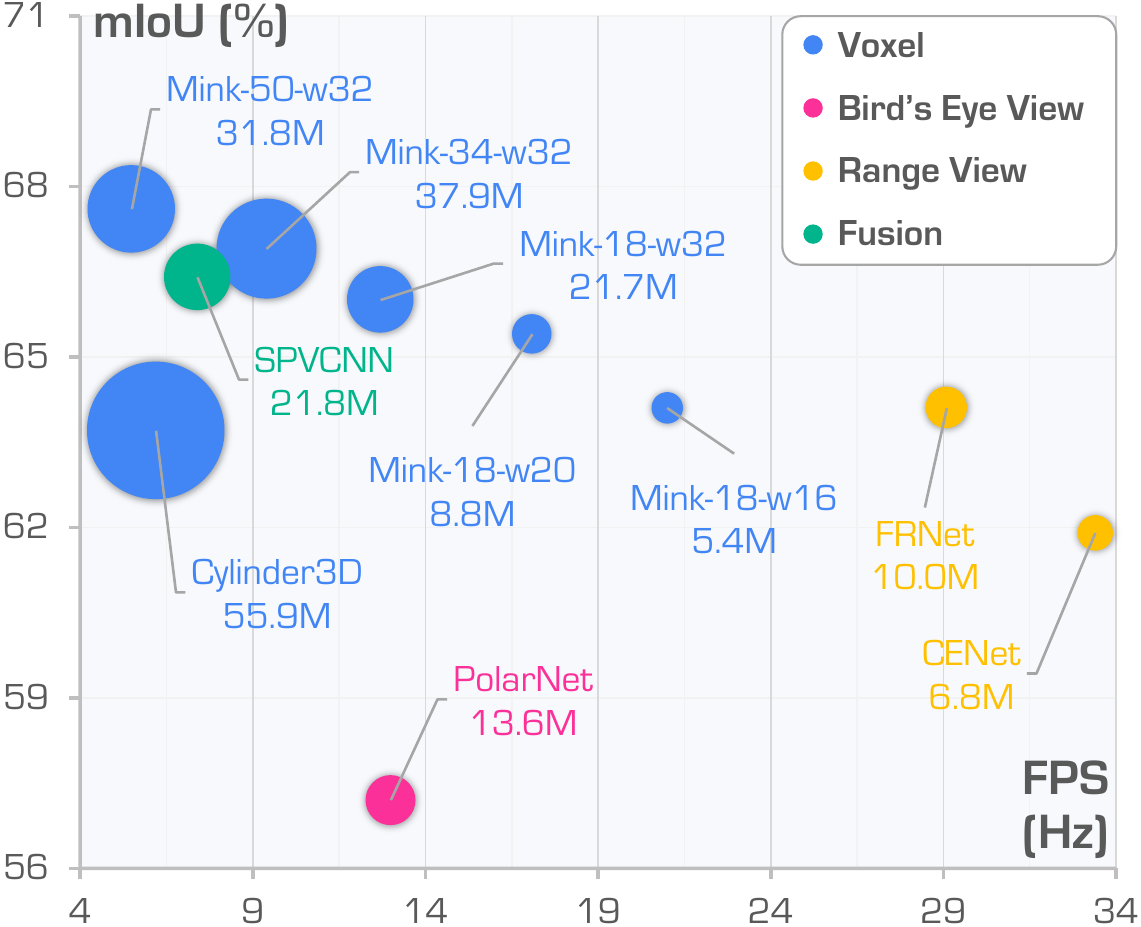}
    \caption{Performance comparisons of state-of-the-art LiDAR segmentation models~\cite{choy2019minkowski,zhu2021cylindrical,tang2020searching,xu2023frnet,cheng2022cenet} from different LiDAR representation groups (voxel, bird's eye view, range view, fusion) on the validation set of SemanticKITTI~\cite{behley2019semanticKITTI}. We report the segmentation accuracy (mIoU), inference latency (FPS), and model parameters. The larger the area coverage, the larger the model capacity.}
    \label{fig:mIoU_vs_fps_sk}
\end{figure}

\noindent\textbf{Implementation Details.} 
In our experiments, we employ AdamW~\cite{loshchilov2018decoupled} as the default optimizer. We set the initial learning rate to 0.01 and utilize the OneCycle scheduler~\cite{smith2019super} to dynamically adjust the learning rate during training. Training is performed across eight A100 GPUs for 50 epochs on SemanticKITTI~\cite{behley2019semanticKITTI} and ScribbleKITTI~\cite{unal2022scribbleKITTI} datasets. For the nuScenes~\cite{fong2022panoptic-nuScenes} dataset, we extend the training to 80 epochs to ensure optimal performance. Each GPU operates with a batch size of 2, making efficient use of parallel processing capabilities. Our default data augmentation pipeline includes point cloud rotation, translation, scaling, and flipping on the global view. Additionally, we introduce a randomized choice of LaserMix~\cite{kong2022laserMix} and PolarMix~\cite{xiao2022polarmix} for the voxel-based, fusion-based, and bird's eye view models, while adopting FrustumMix~\cite{xu2023frnet} for range-view models to achieve better performance. For detailed configurations and reproducibility, please refer to our GitHub repository\footnote{\url{https://github.com/open-mmlab/mmdetection3d}}.

\noindent\textbf{Evaluation Metrics.}
For performance evaluation, we adhere to standard protocols by employing mean Intersection-over-Union (mIoU) as the primary metric and mean Accuracy (mAcc) as a supplementary metric. In addition to these metrics, We evaluate the real-time processing applicability of models by measuring their inference speed. We quantify this using two common metrics: Frames Per Second (FPS) and Iteration Per Second (Iter/s). By incorporating these metrics into our evaluation framework, we ensure a comprehensive assessment of model performance, encompassing both accuracy and efficiency considerations.

\noindent\textbf{Evaluation Protocol.}
To uphold fair evaluation practices in our comparisons, we employ consistent evaluation protocols across different LiDAR segmentation models and data augmentation techniques. One crucial aspect of our methodology is reporting evaluation metrics at the point level rather than voxels or range images. This approach ensures fair comparisons even when models use different rasterizations, as it directly assesses segmentation accuracy at the finest granularity of the input data. Furthermore, we conduct ablation studies to isolate the effects of individual components and augmentations. By systematically varying and analyzing each factor independently, we gain a deeper understanding of its impact on model performance.

\subsection{Benchmark Experiments}
\noindent\textbf{Fully-Supervised LiDAR Semantic Segmentation.} 
We benchmark benchmarking on voxel-based, fusion-based, bird's eye view, and range view segmentors using the validation sets of SemanticKITTI~\cite{behley2019semanticKITTI} and nuScenes~\cite{fong2022panoptic-nuScenes}. The evaluated LiDAR segmentation models include MinkUNet~\cite{choy2019minkowski}, SPVCNN~\cite{tang2020searching}, Cylinder3D~\cite{zhu2021cylindrical}, PolarNet~\cite{zhou2020polarNet}, CENet~\cite{cheng2022cenet}, and FRNet~\cite{xu2023frnet}, as detailed in \cref{tab:main_results}. We report performance under default data augmentation, mixing data augmentation, and, where applicable, test-time augmentation (TTA) for each model. MinkUNet~\cite{choy2019minkowski} achieves the highest mIoU of $71.8\%$ on SemanticKITTI~\cite{behley2019semanticKITTI}, while Cylinder3D~\cite{zhu2021cylindrical} attains $80.9\%$ mIoU on nuScenes~\cite{fong2022panoptic-nuScenes}. Our results in \cref{tab:main_results} illustrate that our default training settings and advanced data augmentation techniques significantly enhance model performance. For instance, MinkUNet's mIoU on SemanticKITTI improved by $5.0\%$ with our default settings, and an additional $3.5\%$ with mixing data augmentation. TTA further boosted performance by $1.4\%$, resulting in nearly a $10\%$ overall improvement. These findings underscore the importance of comprehensive training strategies and robust data augmentation in maximizing model effectiveness.

\noindent\textbf{Inference Speed.}
\cref{fig:mIoU_vs_fps_sk} illustrates the inference speed of each model on the SemanticKITTI~\cite{behley2019semanticKITTI} dataset. All FPS measurements are conducted on a single NVIDIA GeForce RTX 3060 GPU with the data type \texttt{float32}. For fusion- and voxel-based segmentors, SPVCNN~\cite{tang2020searching} employs Torchsparse as its sparse convolution backend, Cylinder3D~\cite{zhu2021cylindrical} uses SpConv, and all variants of MinkUNet~\cite{choy2019minkowski} utilize SpConv v2 as their sparse convolution backend, respectively. The results indicate that range view methods exhibit faster inference speeds, with CENet~\cite{cheng2022cenet} achieving $33$ FPS. While voxel-based methods generally offer higher accuracy, their larger model sizes result in slower inference speeds. Among the MinkUNet variants, MinkUNet-50-w32 achieves the highest accuracy but the slowest inference, whereas MinkUNet-18-w16 offers faster inference with lower accuracy. MinkUNet-34-w32 provides a balanced trade-off between performance and speed, making it a versatile choice for various applications. We believe these insights could become critical for researchers and practitioners who need to optimize for either speed or accuracy depending on the specific requirements of their deployment scenarios, such as real-time, in-vehicle driving scene understanding and segmentation.

\begin{table}[t]
    \centering
    \caption{Results of state-of-the-art LiDAR segmentation models~\cite{choy2019minkowski,zhu2021cylindrical,zhou2020polarNet,cheng2022cenet,xu2023frnet,tang2020searching} under weakly-supervised learning setups on ScribbleKITTI~\cite{unal2022scribbleKITTI}. \textbf{Mix} denote a combination of LaserMix~\cite{kong2022laserMix} and PolarMix~\cite{xiao2022polarmix} or FrustumMix~\cite{xu2023frnet}. \textbf{TTA} denotes test time augmentation. All scores are given in percentage (\%). The \textbf{best} scores under each metric are highlighted in \textbf{bold}.}
    \scalebox{0.92}{
    \begin{tabular}{p{70pt}<{\centering}|p{27pt}<{\centering}p{27pt}<{\centering}|p{35pt}<{\centering}p{35pt}<{\centering}}
        \toprule
        \multirow{2}{*}{\textbf{Model}} & \multirow{2}{*}{\textbf{Mix}} & \multirow{2}{*}{\textbf{TTA}}  &  \multicolumn{2}{c}{\textbf{ScribbleKITTI}} 
        \\
        & & & mIoU & mAcc
        \\\midrule\midrule
        \rowcolor{lblue!9}\multicolumn{5}{l}{\textcolor{lblue}{\textbf{Representation: Voxel}}}
        \\\midrule
        \multirow{3}{*}{MinkUNet} & \textcolor{red}{\ding{55}} & \textcolor{red}{\ding{55}} & $61.2$ & $88.5$
        \\
        & \textcolor{ForestGreen}{\textbf{\checkmark}} & \textcolor{red}{\ding{55}} & $62.6$ & $89.9$ 
        \\
        & \textcolor{ForestGreen}{\textbf{\checkmark}} & \textcolor{ForestGreen}{\textbf{\checkmark}} & $\mathbf{65.7}$ & $\mathbf{90.5}$
        \\\midrule
        \multirow{2}{*}{Cylinder3D} & \textcolor{red}{\ding{55}} & \textcolor{red}{\ding{55}} & $58.8$ & $87.4$ 
        \\
        & \textcolor{ForestGreen}{\textbf{\checkmark}} & \textcolor{red}{\ding{55}} & $59.8$ & $88.8$ 
        \\\midrule
        \rowcolor{lblue!9}\multicolumn{5}{l}{\textcolor{lblue}{\textbf{Representation: Bird's Eye View}}}
        \\\midrule
        \multirow{2}{*}{PolarNet} & \textcolor{red}{\ding{55}} & \textcolor{red}{\ding{55}} & $55.7$ & $87.6$
        \\
        & \textcolor{ForestGreen}{\textbf{\checkmark}} & \textcolor{red}{\ding{55}} & $57.2$ & $88.4$
        \\\midrule
        \rowcolor{lblue!9}\multicolumn{5}{l}{\textcolor{lblue}{\textbf{Representation: Range View}}}
        \\\midrule
        \multirow{3}{*}{FRNet} & \textcolor{red}{\ding{55}} & \textcolor{red}{\ding{55}} & $57.6$ & $88.3$ 
        \\
        & \textcolor{ForestGreen}{\textbf{\checkmark}} & \textcolor{red}{\ding{55}} & $60.2$ & $88.5$ 
        \\
        & \textcolor{ForestGreen}{\textbf{\checkmark}} & \textcolor{ForestGreen}{\textbf{\checkmark}} & $63.1$ & $89.9$
        \\\midrule
        \rowcolor{lblue!9}\multicolumn{5}{l}{\textcolor{lblue}{\textbf{Representation: Fusion}}}
        \\\midrule
        \multirow{3}{*}{SPVCNN} & \textcolor{red}{\ding{55}} & \textcolor{red}{\ding{55}} & $60.1$ & $88.5$
        \\
        & \textcolor{ForestGreen}{\textbf{\checkmark}} & \textcolor{red}{\ding{55}} & $62.9$ & $89.9$
        \\
        & \textcolor{ForestGreen}{\textbf{\checkmark}} & \textcolor{ForestGreen}{\textbf{\checkmark}} & $\mathbf{65.7}$ & $90.4$
        \\
        \bottomrule
    \end{tabular}}
\label{tab:bench_weakly}
\end{table}

\noindent\textbf{Semi-Supervised LiDAR Semantic Segmentation.}
Under the data-efficient learning setting, we leverage partially labeled datasets to evaluate how models can effectively learn with limited supervision. We conduct this study on SemanticKITTI~\cite{behley2019semanticKITTI}, nuScenes~\cite{fong2022panoptic-nuScenes}, and ScribbleKITTI~\cite{unal2022scribbleKITTI}. By using a combination of labeled and unlabeled data, we benchmark methods that employ techniques such as consistency regularization and pseudo-labeling to improve segmentation accuracy, including MeanTeacher~\cite{MeanTeacher}, LaserMix~\cite{kong2022laserMix}, Lim3D~\cite{li23lim3d}, and FrustumMix~\cite{xu2023frnet}. Consistency regularization ensures that predictions remain stable under small perturbations of the input, while pseudo-labeling generates labels for the unlabeled data to iteratively refine the model. Preliminary results, shown in \cref{tab:benchmark_semi}, indicate that semi-supervised learning methods can achieve performance comparable to fully supervised counterparts, demonstrating their potential in reducing the need for extensive manual annotations. These methods not only lower the cost and effort associated with data labeling but also enable the utilization of larger datasets that include both labeled and unlabeled samples, enhancing the robustness and generalization of different LiDAR segmentation approaches.

\noindent\textbf{Weakly-Supervised LiDAR Segmentation.}
For the label-efficient learning task, we utilize the ScribbleKITTI~\cite{unal2022scribbleKITTI} dataset to evaluate LiDAR segmentors trained with weak annotations. This approach significantly reduces the annotation effort by using line scribbles instead of dense labels (which corresponds to around $8.06\%$ of the original labels), providing a practical solution for large-scale dataset creation~\cite{unal2022scribbleKITTI}. Weakly-supervised learning leverages these sparse annotations to guide the model in learning robust features. The performance metrics, presented in \cref{tab:bench_weakly}, reveal that state-of-the-art segmentors~\cite{choy2019minkowski,zhu2021cylindrical,zhou2020polarNet,xu2023frnet,tang2020searching} can attain competitive accuracy with a fraction of the annotation cost, highlighting their efficiency for large-scale applications. These results are particularly promising for applications where detailed annotations are impractical or too costly, suggesting that weak annotations can be a viable alternative for training high-performance models.

\begin{figure*}[t]
    \centering
    \includegraphics[width=\linewidth,scale=1.00]{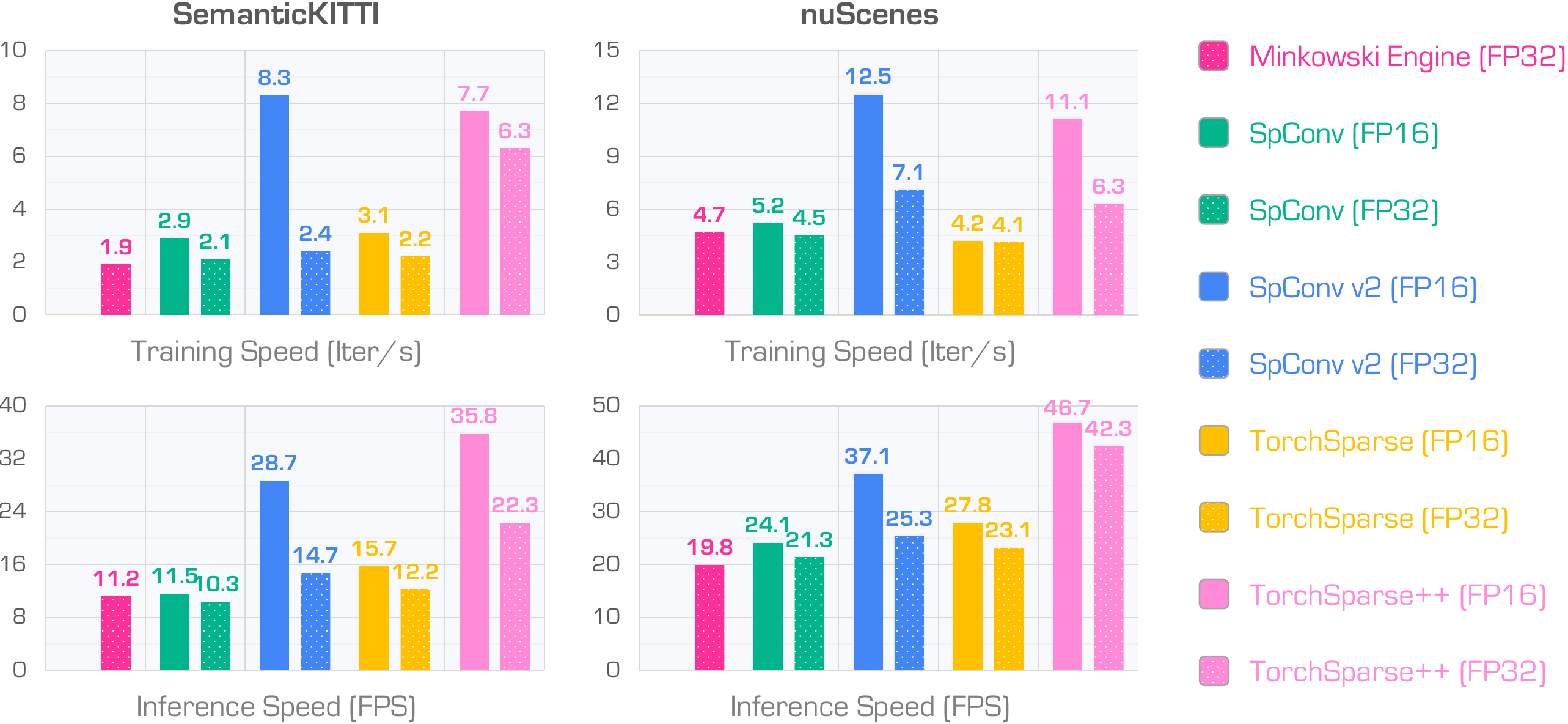}
    \caption{Performance comparisons of different sparse convolution backends~\cite{choy2019minkowski,spconv2022,yan2018second,tang2022torchsparse,tang2023torchsparse++} on the validation sets of SemanticKITTI~\cite{behley2019semanticKITTI} and nuScenes~\cite{fong2022panoptic-nuScenes} datasets. All experiments are conducted using the MinkUNet-34-w32 backbone~\cite{choy2019minkowski}. Both the training speed (Iter/s, iterations per second) and inference speed (FPS, frames per second) are measured using a single NVIDIA A100 GPU.}
    \label{fig:backend}
\end{figure*}

\subsection{Ablation Study}

\noindent\textbf{Sparse Convolution Backends.} 
One of the key features of our MMDetection3D-lidarseg codebase is its support for five common sparse convolution backends: SpConv~\cite{yan2018second}, Minkowski Engine~\cite{choy2019minkowski}, Torchsparse~\cite{tang2022torchsparse}, SpConv v2~\cite{spconv2022}, and Torchsparse++\cite{tang2023torchsparse++}. We compare the efficacy of these sparse convolution backends using the MinkUNet-34-w32 backbone~\cite{choy2019minkowski} on the validation sets of SemanticKITTI~\cite{behley2019semanticKITTI} and nuScenes~\cite{fong2022panoptic-nuScenes}. \cref{fig:backend} reports the training speed (Iter/s, iterations per second) and inference speed (FPS, frames per second) of different backends, with measurements taken on a single NVIDIA A100 GPU. Experiments show that, for all sparse convolution backends, enabling Automatic Mixed Precision (AMP) reduces memory footprint and increases both training iteration speed and inference speed, except for the Minkowski Engine~\cite{choy2019minkowski}, which is not compatible with the \texttt{float16} data type. In terms of training speed, SpConv v2 \cite{spconv2022} with AMP turned on is the fastest, reaching $8.3$ iterations per second on SemanticKITTI and $12.5$ iterations per second on nuScenes. For inference speed, Torchsparse++~\cite{tang2023torchsparse++} using \texttt{float16} data type leads, achieving $35.8$ FPS on SemanticKITTI and $46.7$ FPS on nuScenes. The results also show that all sparse convolutional backends perform faster on nuScenes than on SemanticKITTI due to the denser point clouds in SemanticKITTI, which require higher resolution during voxelization. These findings provide valuable insights into the trade-offs and benefits of each backend, guiding researchers in selecting the appropriate backend based on their specific requirements for training efficiency, memory usage, and inference speed.

\begin{table}[t]
    \centering
    \caption{Ablation study on model capacities. We benchmark six variants of the MinkUNet~\cite{choy2019minkowski} on SemanticKITTI~\cite{behley2019semanticKITTI}. \textbf{Param:} the number of trainable parameters. \textbf{AMP:} whether to use automatic mixed precision during training. \textbf{Mem:} memory consumption. \textbf{Train:} the number of training interactions per second. \textbf{Infer:} the frame-per-second rate during inference. The \textbf{best} scores in terms of training speed (Iter/s), inference speed (FPS), and segmentation accuracy (mIoU) are highlighted in \textbf{bold}.}
    \scalebox{0.92}{
    \begin{tabular}{p{30pt}<{\centering}|p{30pt}<{\centering}|p{30pt}<{\centering}|p{30pt}<{\centering}|p{30pt}<{\centering}|p{30pt}<{\centering}}
        \toprule
        \textbf{Param} (M) & \multirow{2}{*}{\textbf{Amp}} & \textbf{Mem} (GB) & \textbf{Train} (Iter/s) & \textbf{Infer} (FPS) & \textbf{mIoU} (\%)
        \\\midrule\midrule
        \rowcolor{lblue!9}\multicolumn{6}{l}{\textcolor{lblue}{\textbf{Variant: MinkUNet-18-W16}}}
        \\\midrule
        $5.4$ & \textcolor{red}{\ding{55}} & $1.9$ & $7.1$ & $30.7$ & $64.1$ 
        \\
        $5.4$ & \textcolor{ForestGreen}{\textbf{\checkmark}} & $1.1$ & $\mathbf{10.0}$ & $\mathbf{32.4}$ & $64.8$ 
        \\\midrule
        \rowcolor{lblue!9}\multicolumn{6}{l}{\textcolor{lblue}{\textbf{Variant: MinkUNet-18-W20}}}
        \\\midrule
        $8.8$ & \textcolor{red}{\ding{55}} & $2.4$ & $5.3$ & $25.0$ & $65.2$ 
        \\
        $8.8$ & \textcolor{ForestGreen}{\textbf{\checkmark}} & $1.4$ & $8.3$ & $30.3$ & $65.4$ 
        \\\midrule
        \rowcolor{lblue!9}\multicolumn{6}{l}{\textcolor{lblue}{\textbf{Variant: MinkUNet-18-W32}}}
        \\\midrule 
        $21.7$ & \textcolor{red}{\ding{55}} & $3.9$ & $8.3$ & $18.5$ & $66.0$ 
        \\
        $21.7$ & \textcolor{ForestGreen}{\textbf{\checkmark}} & $2.3$ & $\mathbf{10.0}$ & $31.8$ & $65.8$ 
        \\\midrule
        \rowcolor{lblue!9}\multicolumn{6}{l}{\textcolor{lblue}{\textbf{Variant: MinkUNet-34-W32}}}
        \\\midrule
        $37.9$ & \textcolor{red}{\ding{55}} & $4.7$ & $2.4$ & $14.7$ & $66.9$ 
        \\
        37.9 & \textcolor{ForestGreen}{\textbf{\checkmark}} & $2.8$ & $8.3$ & $28.7$ & $66.6$ 
        \\\midrule
        \rowcolor{lblue!9}\multicolumn{6}{l}{\textcolor{lblue}{\textbf{Variant: MinkUNet-50-W32}}}
        \\\midrule
        $31.8$ & \textcolor{red}{\ding{55}} & $11.5$ & $2.8$ & $11.0$ & $65.5 $
        \\
        $31.8$ & \textcolor{ForestGreen}{\textbf{\checkmark}} &  $6.0$ & $7.7$ & $13.7$ & $\mathbf{67.6}$ 
        \\\midrule
        \rowcolor{lblue!9}\multicolumn{6}{l}{\textcolor{lblue}{\textbf{Variant: MinkUNet-101-W32}}}
        \\\midrule
        $70.8$ & \textcolor{red}{\ding{55}} & $14.2$ & $1.9$ & $8.0$ & $66.9$ 
        \\
        $70.8$ & \textcolor{ForestGreen}{\textbf{\checkmark}} & $8.8$ & $5.6$ & $9.9$ & $67.2$ 
        \\
        \bottomrule
    \end{tabular}}
    \label{tab:scale}
\end{table}
\begin{table}[t]
    \centering
    \caption{Ablation study on the use of 3D data augmentation techniques. We benchmark PolarMix~\cite{xiao2022polarmix}, LaserMix~\cite{kong2022laserMix}, and FrustumMix~\cite{xu2023frnet} with the MinkUNet~\cite{choy2019minkowski}, Cylinder3D~\cite{zhu2021cylindrical}, and FRNet~\cite{xu2023frnet} backbones, respectively. The \textbf{best} score under each backbone configuration is highlighted in \textbf{bold}.}
    \scalebox{0.92}{
    \begin{tabular}{p{57pt}<{\centering}|p{57pt}<{\centering}|p{57pt}<{\centering}|p{33pt}<{\centering}}
        \toprule
        \textbf{PolarMix} & \textbf{LaserMix} & \textbf{FrustumMix} & \textbf{mIoU}
        \\
        \cite{xiao2022polarmix} & \cite{kong2022laserMix} & \cite{xu2023frnet} & (\%)
        \\\midrule\midrule
        \rowcolor{lblue!9}\multicolumn{4}{l}{\textcolor{lblue}{\textbf{Model: MinkUNet}}}
        \\\midrule
        \textcolor{red}{\ding{55}} & \textcolor{red}{\ding{55}} & \textcolor{red}{\ding{55}} & $66.9$ 
        \\
        \textcolor{ForestGreen}{\textbf{\checkmark}} & \textcolor{red}{\ding{55}} & \textcolor{red}{\ding{55}} & $68.9$ 
        \\
        \textcolor{red}{\ding{55}} & \textcolor{ForestGreen}{\textbf{\checkmark}} & \textcolor{red}{\ding{55}} & $67.7$ 
        \\
        \textcolor{ForestGreen}{\textbf{\checkmark}} & \textcolor{ForestGreen}{\textbf{\checkmark}} & \textcolor{red}{\ding{55}} & $\mathbf{70.4}$  
        \\\midrule
        \rowcolor{lblue!9}\multicolumn{4}{l}{\textcolor{lblue}{\textbf{Model: Cylinder3D}}}
        \\\midrule
        \textcolor{red}{\ding{55}} & \textcolor{red}{\ding{55}} & \textcolor{red}{\ding{55}} & $63.7$ 
        \\
        \textcolor{ForestGreen}{\textbf{\checkmark}} & \textcolor{red}{\ding{55}} & \textcolor{red}{\ding{55}} & $64.5$ 
        \\
        \textcolor{red}{\ding{55}} & \textcolor{ForestGreen}{\textbf{\checkmark}} & \textcolor{red}{\ding{55}} & $65.6$ 
        \\
        \textcolor{ForestGreen}{\textbf{\checkmark}} & \textcolor{ForestGreen}{\textbf{\checkmark}} & \textcolor{red}{\ding{55}} & $\mathbf{67.0}$
        \\\midrule
        \rowcolor{lblue!9}\multicolumn{4}{l}{\textcolor{lblue}{\textbf{Model: FRNet}}}
        \\\midrule
        \textcolor{red}{\ding{55}} & \textcolor{red}{\ding{55}} & \textcolor{red}{\ding{55}} & $64.1$ 
        \\
        \textcolor{ForestGreen}{\textbf{\checkmark}} & \textcolor{red}{\ding{55}} & \textcolor{red}{\ding{55}} & $66.5$ 
        \\
        \textcolor{red}{\ding{55}} & \textcolor{ForestGreen}{\textbf{\checkmark}} & \textcolor{red}{\ding{55}} & $65.1$ 
        \\
        \textcolor{red}{\ding{55}} & \textcolor{red}{\ding{55}} & \textcolor{ForestGreen}{\textbf{\checkmark}} & $\mathbf{67.6}$ 
        \\
        \bottomrule
    \end{tabular}
    }
\label{tab:mix}
\end{table}

\begin{table}[t]
    \centering
    \caption{Ablation study on the use of the test time augmentation (TTA) during the model inference stage. We benchmark different combinations with the MinkUNet~\cite{choy2019minkowski}, Cylinder3D~\cite{zhu2021cylindrical}, and FRNet~\cite{xu2023frnet} backbones, respectively, on SemanticKITTI~\cite{behley2019semanticKITTI}. The \textbf{best} score under each backbone configuration is highlighted in \textbf{bold}.}
    \scalebox{0.92}{
    \begin{tabular}{p{31pt}<{\centering}|p{31pt}<{\centering}|p{31pt}<{\centering}|p{40pt}<{\centering}|c|r}
        \toprule
        \multirow{2}{*}{\textbf{Flip}} & \multirow{2}{*}{\textbf{Rotate}} & \multirow{2}{*}{\textbf{Scale}} & \multirow{2}{*}{\textbf{Translate}} & \textbf{mIoU} & \multirow{2}{*}{\textbf{Time}}
        \\
        & & & & (\%) & 
        \\
        \midrule\midrule
        \rowcolor{lblue!9}\multicolumn{6}{l}{\textcolor{lblue}{\textbf{Model: MinkUNet}}}
        \\\midrule
        \textcolor{red}{\ding{55}} & \textcolor{red}{\ding{55}} & \textcolor{red}{\ding{55}} & \textcolor{red}{\ding{55}} & $70.4$ & $1\times$
        \\
        \textcolor{ForestGreen}{\textbf{\checkmark}} & \textcolor{red}{\ding{55}} & \textcolor{red}{\ding{55}} & \textcolor{red}{\ding{55}} & $71.2$ & $4\times$
        \\
        \textcolor{ForestGreen}{\textbf{\checkmark}} & \textcolor{ForestGreen}{\textbf{\checkmark}} & \textcolor{red}{\ding{55}} & \textcolor{red}{\ding{55}} & $71.4$ & $12\times$
        \\
        \textcolor{ForestGreen}{\textbf{\checkmark}} & \textcolor{ForestGreen}{\textbf{\checkmark}} & \textcolor{ForestGreen}{\textbf{\checkmark}} & \textcolor{red}{\ding{55}} & $71.7$ & $36\times$ 
        \\
        \textcolor{ForestGreen}{\textbf{\checkmark}} & \textcolor{ForestGreen}{\textbf{\checkmark}} & \textcolor{ForestGreen}{\textbf{\checkmark}} & \textcolor{ForestGreen}{\textbf{\checkmark}} & $\mathbf{71.8}$ & $108\times$
        \\\midrule
        \rowcolor{lblue!9}\multicolumn{6}{l}{\textcolor{lblue}{\textbf{Model: Cylinder3D}}}
        \\\midrule
        \textcolor{red}{\ding{55}} & \textcolor{red}{\ding{55}} & \textcolor{red}{\ding{55}} & \textcolor{red}{\ding{55}} & $67.0$ & $1\times$
        \\
        \textcolor{ForestGreen}{\textbf{\checkmark}} & \textcolor{red}{\ding{55}} & \textcolor{red}{\ding{55}} & \textcolor{red}{\ding{55}} & $65.3$ & $4\times$
        \\
        \textcolor{ForestGreen}{\textbf{\checkmark}} & \textcolor{ForestGreen}{\textbf{\checkmark}} & \textcolor{red}{\ding{55}} & \textcolor{red}{\ding{55}} & $65.4$ & $12\times$
        \\
        \textcolor{ForestGreen}{\textbf{\checkmark}} & \textcolor{ForestGreen}{\textbf{\checkmark}} & \textcolor{ForestGreen}{\textbf{\checkmark}} & \textcolor{red}{\ding{55}} & $69.0$ & $36\times$
        \\
        \textcolor{ForestGreen}{\textbf{\checkmark}} & \textcolor{ForestGreen}{\textbf{\checkmark}} & \textcolor{ForestGreen}{\textbf{\checkmark}} & \textcolor{ForestGreen}{\textbf{\checkmark}} & $\mathbf{69.4}$ & $108\times$
        \\
        \midrule
        \rowcolor{lblue!9}\multicolumn{6}{l}{\textcolor{lblue}{\textbf{Model: FRNet}}}
        \\\midrule
        \textcolor{red}{\ding{55}} & \textcolor{red}{\ding{55}} & \textcolor{red}{\ding{55}} & \textcolor{red}{\ding{55}} & $67.6$ & $1\times$
        \\
        \textcolor{ForestGreen}{\textbf{\checkmark}} & \textcolor{red}{\ding{55}} & \textcolor{red}{\ding{55}} & \textcolor{red}{\ding{55}} & $68.1$ & $4\times$
        \\
        \textcolor{ForestGreen}{\textbf{\checkmark}} & \textcolor{ForestGreen}{\textbf{\checkmark}} & \textcolor{red}{\ding{55}} & \textcolor{red}{\ding{55}} & $68.1$ & $12\times$
        \\
        \textcolor{ForestGreen}{\textbf{\checkmark}} & \textcolor{ForestGreen}{\textbf{\checkmark}} & \textcolor{ForestGreen}{\textbf{\checkmark}} & \textcolor{red}{\ding{55}} & $68.4$ & $36\times$
        \\
        \textcolor{ForestGreen}{\textbf{\checkmark}} & \textcolor{ForestGreen}{\textbf{\checkmark}} & \textcolor{ForestGreen}{\textbf{\checkmark}} & \textcolor{ForestGreen}{\textbf{\checkmark}} & $\mathbf{69.0}$ & $108\times$
        \\
        \bottomrule
    \end{tabular}
    }
    \label{tab:tta}
\end{table}

\noindent\textbf{LiDAR Segmentation Model Capacities.}
We compare six variants of the MinkUNet~\cite{choy2019minkowski} segmentor, evaluating their training memory footprint, training speed, accuracy, and inference speed on SemanticKITTI, as shown in \cref{tab:scale}. Using an NVIDIA A100 GPU for measurements, we find that MinkUNet-50-w32 with mixed precision training achieves the highest mIoU of $67.6\%$, while MinkUNet-18-w16 offers the fastest inference speed at $32.4$ FPS. The commonly used MinkUNet-34-w32 provides an optimal balance between the LiDAR segmentation accuracy and speed, making it a preferred choice for various applications. This detailed comparison allows researchers to make informed decisions about which model variant to use based on their specific needs. MinkUNet-18-w16 and MinkUNet-34-w32 are composed of BasicBlocks, while MinkUNet-50-w32 and MinkUNet-101-w32 are built with Bottleneck Blocks, similar to ResNet architectures. This variety in model complexity and performance metrics provides flexibility for deployment in diverse scenarios, where either high accuracy or faster inference speeds may be prioritized.

\noindent\textbf{Mixing-Based Data Augmentations.}
We evaluate the effectiveness of mixing-based data augmentation techniques on SemanticKITTI~\cite{behley2019semanticKITTI}, with results detailed in \cref{tab:mix}. Using PolarMix~\cite{xiao2022polarmix} and LaserMix~\cite{kong2022laserMix} individually enhances the performance of MinkUNet \cite{choy2019minkowski} by $2\%$ mIoU and $1.1\%$ mIoU, respectively. Combining PolarMix and LaserMix with a random choice strategy yields a $3.5\%$ improvement over using them separately. Similar improvements are observed with Cylinder3D~\cite{zhu2021cylindrical}. For range view segmentors like FRNet~\cite{xu2023frnet}, FrustumMix offers superior performance gains compared to PolarMix or LaserMix, establishing it as the default setting for range view segmentation benchmarks. These results underscore the importance of advanced data augmentation techniques in enhancing model robustness and performance, providing a clear path for further improvements in LiDAR segmentation models. The combination of multiple data augmentation strategies maximizes the exposure of the model to diverse scenarios, improving its ability to generalize and perform well in real-world dense scene understanding applications.

\noindent\textbf{Test Time Augmentation (TTA).}
\cref{tab:tta} presents the impact of various TTA strategies on model performance. Starting with flipping along different axes, TTA improves the performance of MinkUNet~\cite{choy2019minkowski} by $1.6\%$ mIoU. Adding rotation, scaling, and translation further boosts MinkUNet’s performance to $71.8\%$ mIoU on SemanticKITTI~\cite{behley2019semanticKITTI}. While TTA also benefits Cylinder3D~\cite{zhu2021cylindrical} and FRNet~\cite{xu2023frnet}, extensive use of TTA significantly increases inference time. For example, comprehensive TTA techniques result in a $108$-fold increase in inference time compared to no TTA, underscoring the trade-off between performance gains and computational cost. These findings highlight the potential of TTA in improving model accuracy in controlled environments while also illustrating the practical limitations in real-time applications, such as in-vehicle LiDAR segmentation. The choice of TTA techniques should be carefully considered based on the specific requirements of the application, balancing the need for improved accuracy with the constraints of computational resources and inference speed.
\section{Discussions \& Future Directions}
The comprehensive evaluation of LiDAR segmentation models using the MMDetection3D-lidarseg toolbox has provided several key insights into the performance and utility of various models and techniques in real-world applications. Our experiments have highlighted the strengths and weaknesses of different segmentation strategies, data augmentation techniques, and sparse convolution backends, contributing valuable knowledge to the field of LiDAR segmentation. The key takeaways are summarized as follows:
\begin{itemize}
    \item Our codebase unified framework streamlines development and benchmarking by supporting diverse models and backends. This reduces fragmentation and sets new standards for model evaluation.

    \item Advanced 3D data augmentation techniques, such as LaserMix, PolarMix, and FrustumMix significantly enhance model robustness and generalization. Combining these strategies yields notable performance improvements and helps models adapt to diverse environmental conditions.

    \item The comparison of sparse convolution backends shows that SpConv v2 and Torchsparse++ offer superior training and inference speeds with AMP. This is crucial for deploying LiDAR segmentation models in resource-constrained and real-time environments.

    \item Evaluations of MinkUNet variants reveal that larger models like MinkUNet-50-w32 achieve higher accuracy but at increased computational costs, while smaller models offer faster inference at the expense of accuracy. MinkUNet-34-w32 strikes a good balance between accuracy and efficiency.

    \item Test Time Augmentation (TTA) strategies improve model performance but significantly increase inference time, highlighting a trade-off crucial for real-time applications like autonomous driving.
\end{itemize}

Future work will expand MMDetection3D-lidarseg to include more state-of-the-art LiDAR segmentation models, providing a broader range of tools for researchers. Further research into novel data augmentation techniques will aim to enhance model robustness and generalization by simulating a wider variety of real-world scenarios. Optimization of current backends will continue, aiming for greater efficiency and performance. Enhancing the toolbox’s capabilities to support semi-supervised and weakly-supervised learning will reduce the need for extensive manual annotations, facilitating large-scale dataset creation. 

Beyond autonomous driving, the principles and techniques from MMDetection3D-lidarseg can be applied to robotics, augmented reality, and urban planning. Future research will explore these applications, adapting and optimizing the toolbox for various use cases. By making the codebase and trained models publicly available, we aim to foster a collaborative research environment. Contributions from the broader 3D computer vision research community will drive innovation and accelerate the development of more reliable LiDAR segmentation models.

\section{Broader Impact}
\noindent\textbf{Positive Societal Impacts.}
MMDetection3D-lidarseg can significantly advance autonomous driving and related fields by improving LiDAR segmentation. This enhancement can lead to safer and more efficient autonomous systems, reducing traffic accidents and fatalities. Beyond transportation, applications in robotics, augmented reality, urban planning, and environmental monitoring can benefit from improved 3D environment understanding, leading to advancements in automation, immersive experiences, infrastructure development, and accurate environmental mapping.

\noindent\textbf{Negative Societal Impacts.}
The potential negative impacts include the misuse of enhanced LiDAR technology for surveillance, which might raise privacy concerns. Additionally, increased automation could reduce the demand for human labor in industries like transportation and manufacturing, potentially leading to job displacement. To mitigate these risks, it is crucial to adhere to ethical guidelines, ensuring privacy-preserving techniques are integrated into technology development.

\section{Conclusion}
\label{sec:conclusion}

In this work, we introduced MMDetection3D-lidarseg, a comprehensive toolbox designed to facilitate the training and evaluation of state-of-the-art LiDAR segmentation models. By supporting a wide range of segmentation models and integrating advanced data augmentation techniques, our codebase enhances the robustness and generalization capabilities of LiDAR segmentation models. Furthermore, the toolbox provides support for multiple leading sparse convolution backends, optimizing computational efficiency and performance across various hardware configurations.
Our extensive benchmark experiments on widely-used datasets, including SemanticKITTI, nuScenes, and ScribbleKITTI, demonstrate the effectiveness of MMDetection3D-lidarseg in diverse autonomous driving scenarios. The results highlight the toolbox's ability to streamline development, improve benchmarking consistency, and set new standards for research and application in the field of LiDAR segmentation.
By making the codebase and trained models publicly available, MMDetection3D-lidarseg promotes further research, collaboration, and innovation within the community. We believe that this unified framework will accelerate advancements in autonomous driving technologies, ultimately contributing to the development of safer and more reliable autonomous systems.
Future work will focus on expanding the capabilities of MMDetection3D-lidarseg, including the incorporation of additional segmentation models, further optimization of sparse convolution backends, and the integration of more advanced data augmentation techniques. We also plan to explore the application of this toolbox in other domains that can benefit from precise 3D environment understanding, such as robotics and augmented reality.

\section*{Acknowledgements}
We acknowledge the use of the following public resources, during the course of this work:

\begin{itemize}
    \item MMCV\footnote{\url{https://github.com/open-mmlab/mmcv}.} \dotfill Apache License 2.0
    \item MMDetection\footnote{\url{https://github.com/open-mmlab/mmdetection}.} \dotfill Apache License 2.0
    \item MMDetection3D\footnote{\url{https://github.com/open-mmlab/mmdetection3d}.} \dotfill Apache License 2.0
    \item MMEngine\footnote{\url{https://github.com/open-mmlab/mmengine}.} \dotfill Apache License 2.0
    \item OpenPCSeg\footnote{\url{https://github.com/PJLab-ADG/OpenPCSeg}.} \dotfill Apache License 2.0
     \item nuScenes\footnote{\url{https://www.nuscenes.org/nuscenes}.} \dotfill CC BY-NC-SA 4.0
    \item nuScenes-devkit\footnote{\url{https://github.com/nutonomy/nuscenes-devkit}.} \dotfill Apache License 2.0
    \item SemanticKITTI\footnote{\url{http://semantic-kitti.org}.} \dotfill CC BY-NC-SA 4.0
    \item SemanticKITTI-API\footnote{\url{https://github.com/PRBonn/semantic-kitti-api}.} \dotfill MIT License
    \item ScribbleKITTI\footnote{\url{https://github.com/ouenal/scribblekitti}.} \dotfill Unknown
    \item lidar-bonnetal\footnote{\url{https://github.com/PRBonn/lidar-bonnetal}.} \dotfill MIT License
    \item CENet\footnote{\url{https://github.com/huixiancheng/CENet}.} \dotfill MIT License
    \item FRNet\footnote{\url{https://github.com/Xiangxu-0103/FRNet}.} \dotfill Apache License 2.0
    \item PolarSeg\footnote{\url{https://github.com/edwardzhou130/PolarSeg}.} \dotfill BSD 3-Clause License
    \item MinkowskiEngine\footnote{\url{https://github.com/NVIDIA/MinkowskiEngine}.} \dotfill MIT License
    \item TorchSparse\footnote{\url{https://github.com/mit-han-lab/torchsparse}.} \dotfill MIT License
    \item SPVNAS\footnote{\url{https://github.com/mit-han-lab/spvnas}.} \dotfill MIT License
    \item Cylinder3D\footnote{\url{https://github.com/xinge008/Cylinder3D}.} \dotfill Apache License 2.0
    \item SpConv\footnote{\url{https://github.com/traveller59/spconv}.} \dotfill Apache License 2.0
    \item LaserMix\footnote{\url{https://github.com/ldkong1205/LaserMix}.} \dotfill CC BY-NC-SA 4.0
    \item PolarMix\footnote{\url{https://github.com/xiaoaoran/polarmix}.} \dotfill MIT License
\end{itemize}

%%%%%%%%% REFERENCES
{\small
\bibliographystyle{ieee_fullname}
\bibliography{egbib}

\begin{thebibliography}{100}\itemsep=-1pt

\bibitem{ando2023rangevit}
Angelika Ando, Spyros Gidaris, Andrei Bursuc, Gilles Puy, Alexandre Boulch, and Renaud Marlet.
\newblock Rangevit: Towards vision transformers for 3d semantic segmentation in autonomous driving.
\newblock In {\em IEEE/CVF Conference on Computer Vision and Pattern Recognition}, pages 5240--5250, 2023.

\bibitem{aygun2021pls4d}
Mehmet Aygun, Aljosa Osep, Mark Weber, Maxim Maximov, Cyrill Stachniss, Jens Behley, and Laura Leal-Taixé.
\newblock 4d panoptic lidar segmentation.
\newblock In {\em IEEE/CVF Conference on Computer Vision and Pattern Recognition}, pages 5527--5537, 2021.

\bibitem{behley2021semanticKITTI}
Jens Behley, Martin Garbade, Andres Milioto, Jan Quenzel, Sven Behnke, Jürgen Gall, and Cyrill Stachniss.
\newblock Towards 3d lidar-based semantic scene understanding of 3d point cloud sequences: The semantickitti dataset.
\newblock {\em International Journal of Robotics Research}, 40:959--96, 2021.

\bibitem{behley2019semanticKITTI}
Jens Behley, Martin Garbade, Andres Milioto, Jan Quenzel, Sven Behnke, Cyrill Stachniss, and Juergen Gall.
\newblock Semantickitti: A dataset for semantic scene understanding of lidar sequences.
\newblock In {\em IEEE/CVF International Conference on Computer Vision}, pages 9297--9307, 2019.

\bibitem{behley2021lps}
Jens Behley, Andres Milioto, and Cyrill Stachniss.
\newblock A benchmark for lidar-based panoptic segmentation based on kitti.
\newblock In {\em IEEE International Conference on Robotics and Automation}, pages 13596--13603, 2021.

\bibitem{boulch2023also}
Alexandre Boulch, Corentin Sautier, Björn Michele, Gilles Puy, and Renaud Marlet.
\newblock Also: Automotive lidar self-supervision by occupancy estimation.
\newblock In {\em IEEE/CVF Conference on Computer Vision and Pattern Recognition}, pages 13455--13465, 2023.

\bibitem{lang2024pattformer}
Christopher Lang~Alexander Braun, Lars Schillingmann, and Abhinav Valada.
\newblock A point-based approach to efficient lidar multi-task perception.
\newblock {\em arXiv preprint arXiv:2404.12798}, 2024.

\bibitem{caesar2020nuScenes}
Holger Caesar, Varun Bankiti, Alex~H Lang, Sourabh Vora, Venice~Erin Liong, Qiang Xu, Anush Krishnan, Yu Pan, Giancarlo Baldan, and Oscar Beijbom.
\newblock nuscenes: A multimodal dataset for autonomous driving.
\newblock In {\em IEEE/CVF Conference on Computer Vision and Pattern Recognition}, pages 11621--11631, 2020.

\bibitem{cao2022monoscene}
Anh-Quan Cao and Raoul~De Charette.
\newblock Monoscene: Monocular 3d semantic scene completion.
\newblock In {\em IEEE/CVF Conference on Computer Vision and Pattern Recognition}, pages 3991--4001, 2022.

\bibitem{cao2024pasco}
Anh-Quan Cao, Angela Dai, and Raoul de Charette.
\newblock Pasco: Urban 3d panoptic scene completion with uncertainty awareness.
\newblock In {\em IEEE/CVF Conference on Computer Vision and Pattern Recognition}, 2024.

\bibitem{cen2022open}
Jun Cen, Peng Yun, Shiwei Zhang, Junhao Cai, Di Luan, Mingqian Tang, Ming Liu, and Michael~Yu Wang.
\newblock Open-world semantic segmentation for lidar point clouds.
\newblock In {\em European Conference on Computer Vision}, pages 318--334, 2022.

\bibitem{chen2021polarStream}
Qi Chen, Sourabh Vora, and Oscar Beijbom.
\newblock Polarstream: Streaming lidar object detection and segmentation with polar pillars.
\newblock In {\em Advances in Neural Information Processing Systems}, volume~34, 2021.

\bibitem{chen2023towards}
Runnan Chen, Youquan Liu, Lingdong Kong, Nenglun Chen, Xinge Zhu, Yuexin Ma, Tongliang Liu, and Wenping Wang.
\newblock Towards label-free scene understanding by vision foundation models.
\newblock In {\em Advances in Neural Information Processing Systems}, volume~36, 2023.

\bibitem{2023CLIP2Scene}
Runnan Chen, Youquan Liu, Lingdong Kong, Xinge Zhu, Yuexin Ma, Yikang Li, Yuenan Hou, Yu Qiao, and Wenping Wang.
\newblock Clip2scene: Towards label-efficient 3d scene understanding by clip.
\newblock In {\em IEEE/CVF Conference on Computer Vision and Pattern Recognition}, pages 7020--7030, 2023.

\bibitem{chen2021moving}
Xieyuanli Chen, Shijie Li, Benedikt Mersch, Louis Wiesmann, Jürgen Gall, Jens Behley, and Cyrill Stachniss.
\newblock Moving object segmentation in 3d lidar data: A learning-based approach exploiting sequential data.
\newblock {\em IEEE Robotics and Automation Letters}, 6(4):6529--6536, 2021.

\bibitem{CPS}
Xiaokang Chen, Yuhui Yuan, Gang Zeng, and Jingdong Wang.
\newblock Semi-supervised semantic segmentation with cross pseudo supervision.
\newblock In {\em IEEE/CVF Conference on Computer Vision and Pattern Recognition}, pages 2613--2622, 2021.

\bibitem{cheng2022cenet}
Huixian Cheng, Xianfeng Han, and Guoqiang Xiao.
\newblock Cenet: Toward concise and efficient lidar semantic segmentation for autonomous driving.
\newblock In {\em IEEE International Conference on Multimedia and Expo}, pages 1--6, 2022.

\bibitem{cheng2023transrvnet}
Huixian Cheng, Xianfeng Han, and Guoqiang Xiao.
\newblock Transrvnet: Lidar semantic segmentation with transformer.
\newblock {\em IEEE Transactions on Intelligent Transportation Systems}, 24(6):5895--5907, 2023.

\bibitem{cheng2021af2S3Net}
Ran Cheng, Ryan Razani, Ehsan Taghavi, Enxu Li, and Bingbing Liu.
\newblock Af2-s3net: Attentive feature fusion with adaptive feature selection for sparse semantic segmentation network.
\newblock In {\em IEEE/CVF Conference on Computer Vision and Pattern Recognition}, pages 12547--12556, 2021.

\bibitem{choy2019minkowski}
Christopher Choy, JunYoung Gwak, and Silvio Savarese.
\newblock 4d spatio-temporal convnets: Minkowski convolutional neural networks.
\newblock In {\em IEEE/CVF Conference on Computer Vision and Pattern Recognition}, pages 3075--3084, 2019.

\bibitem{mmdet3d2020}
MMDetection3D Contributors.
\newblock {MMDetection3D: OpenMMLab} next-generation platform for general {3D} object detection.
\newblock \url{https://github.com/open-mmlab/mmdetection3d}, 2020.

\bibitem{pointcept2023}
Pointcept Contributors.
\newblock Pointcept: A codebase for point cloud perception research.
\newblock \url{https://github.com/Pointcept/Pointcept}, 2023.

\bibitem{spconv2022}
Spconv Contributors.
\newblock Spconv: Spatially sparse convolution library.
\newblock \url{https://github.com/traveller59/spconv}, 2022.

\bibitem{cortinhal2020salsanext}
Tiago Cortinhal, George Tzelepis, and Eren~Erdal Aksoy.
\newblock Salsanext: Fast, uncertainty-aware semantic segmentation of lidar point clouds.
\newblock In {\em International Symposium on Visual Computing}, pages 207--222, 2020.

\bibitem{douillard201lidarseg}
Bertrand Douillard, James Underwood, Noah Kuntz, Vsevolod Vlaskine, Alastair Quadros, Peter Morton, and Alon Frenkel.
\newblock On the segmentation of 3d lidar point clouds.
\newblock In {\em IEEE International Conference on Robotics and Automation}, pages 2798--2805, 2011.

\bibitem{ester1996dbscan}
Martin Ester, Hans-Peter Kriegel, Jörg Sander, and Xiaowei Xu.
\newblock A density-based algorithm for discovering clusters in large spatial databases with noise.
\newblock In {\em ACM SIGKDD Conference on Knowledge Discovery and Data Mining}, pages 226--231, 1996.

\bibitem{foschler1981ransac}
Martin~A. Fischler and Robert~C. Bolles.
\newblock Random sample consensus: A paradigm for model fitting with applications to image analysis and automated cartography.
\newblock {\em Communications of the ACM}, 24(6):381--395, 1981.

\bibitem{fong2022panoptic-nuScenes}
Whye~Kit Fong, Rohit Mohan, Juana~Valeria Hurtado, Lubing Zhou, Holger Caesar, Oscar Beijbom, and Abhinav Valada.
\newblock Panoptic nuscenes: A large-scale benchmark for lidar panoptic segmentation and tracking.
\newblock {\em IEEE Robotics and Automation Letters}, 7(2):3795--3802, 2022.

\bibitem{CutMix-Seg}
Geoff French, Timo Aila, Samuli Laine, Michal Mackiewicz, and Graham Finlayson.
\newblock Semi-supervised semantic segmentation needs strong, high-dimensional perturbations.
\newblock In {\em British Machine Vision Conference}, 2020.

\bibitem{gao2021survey}
Biao Gao, Yancheng Pan, Chengkun Li, Sibo Geng, and Huijing Zhao.
\newblock Are we hungry for 3d lidar data for semantic segmentation? a survey of datasets and methods.
\newblock {\em IEEE Transactions on Intelligent Transportation Systems}, 23(7):6063--6081, 2021.

\bibitem{gasperini2021panoster}
Stefano Gasperini, Mohammad-Ali~Nikouei Mahani, Alvaro Marcos-Ramiro, Nassir Navab, and Federico Tombari.
\newblock Panoster: End-to-end panoptic segmentation of lidar point clouds.
\newblock {\em IEEE Robotics and Automation Letters}, 6(2):3216--3223, 2021.

\bibitem{geiger2012kitti}
Andreas Geiger, Philip Lenz, and Raquel Urtasun.
\newblock Are we ready for autonomous driving? the kitti vision benchmark suite.
\newblock In {\em IEEE/CVF Conference on Computer Vision and Pattern Recognition}, pages 3354--3361, 2012.

\bibitem{hahner2022snowfall}
Martin Hahner, Christos Sakaridis, Mario Bijelic, Felix Heide, Fisher Yu, Dengxin Dai, and Luc~Van Gool.
\newblock Lidar snowfall simulation for robust 3d object detection.
\newblock In {\em IEEE/CVF Conference on Computer Vision and Pattern Recognition}, pages 16364--16374, 2022.

\bibitem{hahner2021fog}
Martin Hahner, Christos Sakaridis, Dengxin Dai, and Luc~Van Gool.
\newblock Fog simulation on real lidar point clouds for 3d object detection in adverse weather.
\newblock In {\em IEEE/CVF International Conference on Computer Vision}, pages 15283--15292, 2021.

\bibitem{hong20224dDSNet}
Fangzhou Hong, Lingdong Kong, Hui Zhou, Xinge Zhu, Hongsheng Li, and Ziwei Liu.
\newblock Unified 3d and 4d panoptic segmentation via dynamic shifting networks.
\newblock {\em IEEE Transactions on Pattern Analysis and Machine Intelligence}, 46(5):3480--3495, 2024.

\bibitem{pvkd}
Yuenan Hou, Xinge Zhu, Yuexin Ma, Chen~Change Loy, and Yikang Li.
\newblock Point-to-voxel knowledge distillation for lidar semantic segmentation.
\newblock In {\em IEEE Conference on Computer Vision and Pattern Recognition}, pages 8479--8488, 2022.

\bibitem{hu2021sensatUrban}
Qingyong Hu, Bo Yang, Sheikh Khalid, Wen Xiao, Niki Trigoni, and Andrew Markham.
\newblock Towards semantic segmentation of urban-scale 3d point clouds: A dataset, benchmarks and challenges.
\newblock In {\em IEEE/CVF Conference on Computer Vision and Pattern Recognition}, pages 4977--4987, 2021.

\bibitem{hu2020randla}
Qingyong Hu, Bo Yang, Linhai Xie, Stefano Rosa, Yulan Guo, Zhihua Wang, Niki Trigoni, and Andrew Markham.
\newblock Randla-net: Efficient semantic segmentation of large-scale point clouds.
\newblock In {\em IEEE/CVF Conference on Computer Vision and Pattern Recognition}, pages 11108--11117, 2020.

\bibitem{jaritz2020xMUDA}
Maximilian Jaritz, Tuan-Hung Vu, Raoul de Charette, Emilie Wirbel, and Patrick Pérez.
\newblock xmuda: Cross-modal unsupervised domain adaptation for 3d semantic segmentation.
\newblock In {\em IEEE/CVF Conference on Computer Vision and Pattern Recognition}, pages 12605--12614, 2020.

\bibitem{jhaldiyal2023survey}
Alok Jhaldiyal and Navendu Chaudhary.
\newblock Semantic segmentation of 3d lidar data using deep learning: a review of projection-based methods.
\newblock {\em Applied Intelligence}, 53(6):6844--6855, 2023.

\bibitem{kong2023rethinking}
Lingdong Kong, Youquan Liu, Runnan Chen, Yuexin Ma, Xinge Zhu, Yikang Li, Yuenan Hou, Yu Qiao, and Ziwei Liu.
\newblock Rethinking range view representation for lidar segmentation.
\newblock In {\em IEEE/CVF International Conference on Computer Vision}, pages 228--240, 2023.

\bibitem{kong2023robo3D}
Lingdong Kong, Youquan Liu, Xin Li, Runnan Chen, Wenwei Zhang, Jiawei Ren, Liang Pan, Kai Chen, and Ziwei Liu.
\newblock Robo3d: Towards robust and reliable 3d perception against corruptions.
\newblock In {\em IEEE/CVF International Conference on Computer Vision}, pages 19994--20006, 2023.

\bibitem{kong2023conDA}
Lingdong Kong, Niamul Quader, and Venice~Erin Liong.
\newblock Conda: Unsupervised domain adaptation for lidar segmentation via regularized domain concatenation.
\newblock In {\em IEEE International Conference on Robotics and Automation}, pages 9338--9345, 2023.

\bibitem{kong2022laserMix}
Lingdong Kong, Jiawei Ren, Liang Pan, and Ziwei Liu.
\newblock Lasermix for semi-supervised lidar semantic segmentation.
\newblock In {\em IEEE/CVF Conference on Computer Vision and Pattern Recognition}, pages 21705--21715, 2023.

\bibitem{kong2023robodepth}
Lingdong Kong, Shaoyuan Xie, Hanjiang Hu, Lai~Xing Ng, Benoit~R. Cottereau, and Wei~Tsang Ooi.
\newblock Robodepth: Robust out-of-distribution depth estimation under corruptions.
\newblock In {\em Advances in Neural Information Processing Systems}, volume~36, 2023.

\bibitem{kong2024calib3d}
Lingdong Kong, Xiang Xu, Jun Cen, Wenwei Zhang, Liang Pan, Kai Chen, and Ziwei Liu.
\newblock Calib3d: Calibrating model preferences for reliable 3d scene understanding.
\newblock {\em arXiv preprint arXiv:2403.17010}, 2024.

\bibitem{kong2024lasermix2}
Lingdong Kong, Xiang Xu, Jiawei Ren, Wenwei Zhang, Liang Pan, Kai Chen, Wei~Tsang Ooi, and Ziwei Liu.
\newblock Multi-modal data-efficient 3d scene understanding for autonomous driving.
\newblock {\em arXiv preprint arXiv:2405.05258}, 2024.

\bibitem{landrieu2017structured}
Loic Landrieu, Hugo Raguet, Bruno Vallet, Cl{\'e}ment Mallet, and Martin Weinmann.
\newblock A structured regularization framework for spatially smoothing semantic labelings of 3d point clouds.
\newblock {\em ISPRS Journal of Photogrammetry and Remote Sensing}, 132:102--118, 2017.

\bibitem{li2022sdseg3d}
Jiale Li, Hang Dai, and Yong Ding.
\newblock Self-distillation for robust lidar semantic segmentation in autonomous driving.
\newblock In {\em European Conference on Computer Vision}, pages 659--676, 2022.

\bibitem{li2023mseg3d}
Jiale Li, Hang Dai, Hao Han, and Yong Ding.
\newblock Mseg3d: Multi-modal 3d semantic segmentation for autonomous driving.
\newblock In {\em IEEE/CVF Conference on Computer Vision and Pattern Recognition}, pages 21694--21704, 2023.

\bibitem{li23lim3d}
Li Li, Hubert P.~H. Shum, and Toby~P. Breckon.
\newblock Less is more: Reducing task and model complexity for 3d point cloud semantic segmentation.
\newblock In {\em IEEE/CVF Conference on Computer Vision and Pattern Recognition}, pages 9361--9371, 2023.

\bibitem{li2022coarse3D}
Rong Li, Raoul de Charette, and Anh-Quan Cao.
\newblock Coarse3d: Class-prototypes for contrastive learning in weakly-supervised 3d point cloud segmentation.
\newblock In {\em British Machine Vision Conference}, 2022.

\bibitem{li2023tfnet}
Rong Li, Shijie Li, Xieyuanli Chen, Teli Ma, Wang Hao, Juergen Gall, and Junwei Liang.
\newblock Tfnet: Exploiting temporal cues for fast and accurate lidar semantic segmentation.
\newblock {\em arXiv preprint arXiv:2309.07849}, 2023.

\bibitem{li2024place3d}
Ye Li, Lingdong Kong, Hanjiang Hu, Xiaohao Xu, and Xiaonan Huang.
\newblock Optimizing lidar placements for robust driving perception in adverse conditions.
\newblock {\em arXiv preprint arXiv:2403.17009}, 2024.

\bibitem{liao2024vlm2scene}
Guibiao Liao, Jiankun Li, and Xiaoqing Ye.
\newblock Vlm2scene: Self-supervised image-text-lidar learning with foundation models for autonomous driving scene understanding.
\newblock In {\em AAAI Conference on Artificial Intelligence}, pages 3351--3359, 2024.

\bibitem{liong2020amvNet}
Venice~Erin Liong, Thi Ngoc~Tho Nguyen, Sergi Widjaja, Dhananjai Sharma, and Zhuang~Jie Chong.
\newblock Amvnet: Assertion-based multi-view fusion network for lidar semantic segmentation.
\newblock {\em arXiv preprint arXiv:2012.04934}, 2020.

\bibitem{liu2022less}
Minghua Liu, Yin Zhou, Charles~R. Qi, Boqing Gong, Hao Su, and Dragomir Anguelov.
\newblock Less: Label-efficient semantic segmentation for lidar point clouds.
\newblock In {\em European Conference on Computer Vision}, pages 70--89, 2022.

\bibitem{openpcseg2023}
Youquan Liu, Yeqi Bai, Lingdong Kong, Runnan Chen, Yuenan Hou, Botian Shi, and Li Yikang.
\newblock Openpcseg: An open source point cloud segmentation codebase.
\newblock \url{https://github.com/PJLab-ADG/PCSeg}, 2023.

\bibitem{liu2023uniseg}
Youquan Liu, Runnan Chen, Xin Li, Lingdong Kong, Yuchen Yang, Zhaoyang Xia, Yeqi Bai, Xinge Zhu, Yuexin Ma, Yikang Li, Yu Qiao, and Yuenan Hou.
\newblock Uniseg: A unified multi-modal lidar segmentation network and the openpcseg codebase.
\newblock In {\em IEEE/CVF International Conference on Computer Vision}, pages 21662--21673, 2023.

\bibitem{liu2023seal}
Youquan Liu, Lingdong Kong, Jun Cen, Runnan Chen, Wenwei Zhang, Liang Pan, Kai Chen, and Ziwei Liu.
\newblock Segment any point cloud sequences by distilling vision foundation models.
\newblock In {\em Advances in Neural Information Processing Systems}, volume~36, 2023.

\bibitem{liu2024m3net}
Youquan Liu, Lingdong Kong, Xiaoyang Wu, Runnan Chen, Xin Li, Liang Pan, Ziwei Liu, and Yuexin Ma.
\newblock Multi-space alignments towards universal lidar segmentation.
\newblock In {\em IEEE/CVF Conference on Computer Vision and Pattern Recognition}, 2024.

\bibitem{loiseau2022online}
Romain Loiseau, Mathieu Aubry, and Loïc Landrieu.
\newblock Online segmentation of lidar sequences: Dataset and algorithm.
\newblock In {\em European Conference on Computer Vision}, pages 301--317, 2022.

\bibitem{loshchilov2018decoupled}
Ilya Loshchilov and Frank Hutter.
\newblock Decoupled weight decay regularization.
\newblock In {\em International Conference on Learning Representations}, 2018.

\bibitem{marcuzzi2023mask}
Rodrigo Marcuzzi, Lucas Nunes, Louis Wiesmann, Jens Behley, and Cyrill Stachniss.
\newblock Mask-based panoptic lidar segmentation for autonomous driving.
\newblock {\em IEEE Robotics and Automation Letters}, 8(2):1141--1148, 2023.

\bibitem{meng2024small}
Qiang Meng, Xiao Wang, JiaBao Wang, Liujiang Yan, and Ke Wang.
\newblock Small, versatile and mighty: A range-view perception framework.
\newblock {\em arXiv preprint arXiv:2403.00325}, 2024.

\bibitem{michele2024saluda}
Bjoern Michele, Alexandre Boulch, Gilles Puy, Tuan-Hung Vu, Renaud Marlet, and Nicolas Courty.
\newblock Saluda: Surface-based automotive lidar unsupervised domain adaptation.
\newblock In {\em International Conference on 3D Vision}, 2024.

\bibitem{milioto2019rangenet++}
Andres Milioto, Ignacio Vizzo, Jens Behley, and Cyrill Stachniss.
\newblock Rangenet++: Fast and accurate lidar semantic segmentation.
\newblock In {\em IEEE/RSJ International Conference on Intelligent Robots and Systems}, pages 4213--4220, 2019.

\bibitem{osep2024sal}
Aljoša Ošep, Tim Meinhardt, Francesco Ferroni, Neehar Peri, Deva Ramanan, and Laura Leal-Taixé.
\newblock Better call sal: Towards learning to segment anything in lidar.
\newblock {\em arXiv preprint arXiv:2403.13129}, 2024.

\bibitem{piroli2024label}
Aldi Piroli, Vinzenz Dallabetta, Johannes Kopp, Marc Walessa, Daniel Meissner, and Klaus Dietmayer.
\newblock Label-efficient semantic segmentation of lidar point clouds in adverse weather conditions.
\newblock {\em IEEE Robotics and Automation Letters}, 9(6):5575--5582, 2024.

\bibitem{puy23waffleiron}
Gilles Puy, Alexandre Boulch, and Renaud Marlet.
\newblock Using a waffle iron for automotive point cloud semantic segmentation.
\newblock In {\em IEEE/CVF International Conference on Computer Vision}, pages 3379--3389, 2023.

\bibitem{puy2024three}
Gilles Puy, Spyros Gidaris, Alexandre Boulch, Oriane Siméoni, Corentin Sautier, Patrick Pérez, Andrei Bursuc, and Renaud Marlet.
\newblock Three pillars improving vision foundation model distillation for lidar.
\newblock In {\em IEEE/CVF Conference on Computer Vision and Pattern Recognition}, 2024.

\bibitem{rizzoli2022survey}
Giulia Rizzoli, Francesco Barbato, and Pietro Zanuttigh.
\newblock Multimodal semantic segmentation in autonomous driving: A review of current approaches and future perspectives.
\newblock {\em Technologies}, 10(4):90, 2022.

\bibitem{sautier2024bevcontrast}
Corentin Sautier, Gilles Puy, Alexandre Boulch, Renaud Marlet, and Vincent Lepetit.
\newblock Bevcontrast: Self-supervision in bev space for automotive lidar point clouds.
\newblock In {\em International Conference on 3D Vision}, 2024.

\bibitem{sirohi2021efficientlps}
Kshitij Sirohi, Rohit Mohan, Daniel Büscher, Wolfram Burgard, and Abhinav Valada.
\newblock Efficientlps: Efficient lidar panoptic segmentation.
\newblock {\em IEEE Transactions on Robotics}, 38(3):1894--1914, 2021.

\bibitem{smith2019super}
Leslie~N Smith and Nicholay Topin.
\newblock Super-convergence: Very fast training of neural networks using large learning rates.
\newblock In {\em Artificial Intelligence and Machine Learning for Multi-Domain Operations Applications}, volume 11006, pages 369--386, 2019.

\bibitem{sun2020waymoOpen}
Pei Sun, Henrik Kretzschmar, Xerxes Dotiwalla, Aurelien Chouard, Vijaysai Patnaik, Paul Tsui, James Guo, Yin Zhou, Yuning Chai, Benjamin Caine, Vijay Vasudevan, Wei Han, Jiquan Ngiam, Hang Zhao, Aleksei Timofeev, Scott Ettinger, Maxim Krivokon, Amy Gao, Aditya Joshi, Yu Zhang, Jonathon Shlens, Zhifeng Chen, and Dragomir Anguelov.
\newblock Scalability in perception for autonomous driving: Waymo open dataset.
\newblock In {\em IEEE/CVF Conference on Computer Vision and Pattern Recognition}, pages 2446--2454, 2020.

\bibitem{tang2022torchsparse}
Haotian Tang, Zhijian Liu, Xiuyu Li, Yujun Lin, and Song Han.
\newblock Torchsparse: Efficient point cloud inference engine.
\newblock {\em Proceedings of Machine Learning and Systems}, 4:302--315, 2022.

\bibitem{tang2020searching}
Haotian Tang, Zhijian Liu, Shengyu Zhao, Yujun Lin, Ji Lin, Hanrui Wang, and Song Han.
\newblock Searching efficient 3d architectures with sparse point-voxel convolution.
\newblock In {\em European Conference on Computer Vision}, pages 685--702, 2020.

\bibitem{tang2023torchsparse++}
Haotian Tang, Shang Yang, Zhijian Liu, Ke Hong, Zhongming Yu, Xiuyu Li, Guohao Dai, Yu Wang, and Song Han.
\newblock Torchsparse++: Efficient point cloud engine.
\newblock In {\em IEEE/CVF Conference on Computer Vision and Pattern Recognition Workshops}, pages 202--209, 2023.

\bibitem{MeanTeacher}
Antti Tarvainen and Harri Valpola.
\newblock Mean teachers are better role models: Weight-averaged consistency targets improve semi-supervised deep learning results.
\newblock In {\em Advances in Neural Information Processing Systems}, volume~30, 2017.

\bibitem{thomas2019kpconv}
Hugues Thomas, Charles~R Qi, Jean-Emmanuel Deschaud, Beatriz Marcotegui, Fran{\c{c}}ois Goulette, and Leonidas~J Guibas.
\newblock Kpconv: Flexible and deformable convolution for point clouds.
\newblock In {\em IEEE/CVF International Conference on Computer Vision}, pages 6411--6420, 2019.

\bibitem{triess2021survey}
Larissa~T. Triess, Mariella Dreissig, Christoph~B. Rist, and J.~Marius Zöllner.
\newblock A survey on deep domain adaptation for lidar perception.
\newblock In {\em IEEE Intelligent Vehicles Symposium Workshops}, pages 350--357, 2021.

\bibitem{triess2020scan}
Larissa~T. Triess, David Peter, Christoph~B. Rist, and J.~Marius Z{\"o}llner.
\newblock Scan-based semantic segmentation of lidar point clouds: An experimental study.
\newblock In {\em IEEE Intelligent Vehicles Symposium}, pages 1116--1121, 2020.

\bibitem{uecker2022analyzing}
Marc Uecker, Tobias Fleck, Marcel Pflugfelder, and J.~Marius Zöllner.
\newblock Analyzing deep learning representations of point clouds for real-time in-vehicle lidar perception.
\newblock {\em arXiv preprint arXiv:2210.14612}, 2022.

\bibitem{unal2022scribbleKITTI}
Ozan Unal, Dengxin Dai, and Luc~Van Gool.
\newblock Scribble-supervised lidar semantic segmentation.
\newblock In {\em IEEE/CVF Conference on Computer Vision and Pattern Recognition}, pages 2697--2707, 2022.

\bibitem{varney2020dales}
Nina Varney, Vijayan~K. Asari, and Quinn Graehling.
\newblock Dales: A large-scale aerial lidar data set for semantic segmentation.
\newblock In {\em IEEE/CVF Conference on Computer Vision and Pattern Recognition Workshops}, pages 186--187, 2020.

\bibitem{wang2018pointseg}
Yuan Wang, Tianyue Shi, Peng Yun, Lei Tai, and Ming Liu.
\newblock Pointseg: Real-time semantic segmentation based on 3d lidar point cloud.
\newblock {\em arXiv preprint arXiv:1807.06288}, 2018.

\bibitem{weinmann2015semantic}
Martin Weinmann, Boris Jutzi, Stefan Hinz, and Cl{\'e}ment Mallet.
\newblock Semantic point cloud interpretation based on optimal neighborhoods, relevant features and efficient classifiers.
\newblock {\em ISPRS Journal of Photogrammetry and Remote Sensing}, 105:286--304, 2015.

\bibitem{xiao2022polarmix}
Aoran Xiao, Jiaxing Huang, Dayan Guan, Kaiwen Cui, Shijian Lu, and Ling Shao.
\newblock Polarmix: A general data augmentation technique for lidar point clouds.
\newblock In {\em Advances in Neural Information Processing Systems}, volume~35, pages 11035--11048, 2022.

\bibitem{xiao2022transfer}
Aoran Xiao, Jiaxing Huang, Dayan Guan, Fangneng Zhan, and Shijian Lu.
\newblock Transfer learning from synthetic to real lidar point cloud for semantic segmentation.
\newblock In {\em AAAI Conference on Artificial Intelligence}, pages 12795--2803, 2022.

\bibitem{xiao2023survey}
Aoran Xiao, Jiaxing Huang, Dayan Guan, Xiaoqin Zhang, Shijian Lu, and Ling Shao.
\newblock Unsupervised point cloud representation learning with deep neural networks: A survey.
\newblock {\em IEEE Transactions on Pattern Analysis and Machine Intelligence}, 45(9):11321--11339, 2023.

\bibitem{xie2023robobev}
Shaoyuan Xie, Lingdong Kong, Wenwei Zhang, Jiawei Ren, Liang Pan, Kai Chen, and Ziwei Liu.
\newblock Robobev: Towards robust bird's eye view perception under corruptions.
\newblock {\em arXiv preprint arXiv:2304.06719}, 2023.

\bibitem{xu2020squeezesegv3}
Chenfeng Xu, Bichen Wu, Zining Wang, Wei Zhan, Peter Vajda, Kurt Keutzer, and Masayoshi Tomizuka.
\newblock Squeezesegv3: Spatially-adaptive convolution for efficient point-cloud segmentation.
\newblock In {\em European Conference on Computer Vision}, pages 1--19, 2020.

\bibitem{xu2024visual}
Jingyi Xu, Weidong Yang, Lingdong Kong, Youquan Liu, Rui Zhang, Qingyuan Zhou, and Ben Fei.
\newblock Visual foundation models boost cross-modal unsupervised domain adaptation for 3d semantic segmentation.
\newblock {\em arXiv preprint arXiv:2403.10001}, 2024.

\bibitem{xu2023frnet}
Xiang Xu, Lingdong Kong, Hui Shuai, and Qingshan Liu.
\newblock Frnet: Frustum-range networks for scalable lidar segmentation.
\newblock {\em arXiv preprint arXiv:2312.04484}, 2023.

\bibitem{yan2022dpass}
Xu Yan, Jiantao Gao, Chao~Zheng Chaoda~Zheng, Ruimao Zhang, Shuguang Cui, and Zhen Li.
\newblock 2dpass: 2d priors assisted semantic segmentation on lidar point clouds.
\newblock In {\em European Conference on Computer Vision}, pages 677--695, 2022.

\bibitem{yan2018second}
Yan Yan, Yuxing Mao, and Bo Li.
\newblock Second: Sparsely embedded convolutional detection.
\newblock {\em Sensors}, 18(10):3337, 2018.

\bibitem{yu2022benchmarking}
Kaicheng Yu, Tang Tao, Hongwei Xie, Zhiwei Lin, Zhongwei Wu, Zhongyu Xia, Tingting Liang, Haiyang Sun, Jiong Deng, Dayang Hao, Yongtao Wang, Xiaodan Liang, and Bing Wang.
\newblock Benchmarking the robustness of lidar-camera fusion for 3d object detection.
\newblock {\em arXiv preprint arXiv:2205.14951}, 2022.

\bibitem{zermas2017fast}
Dimitris Zermas, Izzat Izzat, and Nikolaos Papanikolopoulos.
\newblock Fast segmentation of 3d point clouds: A paradigm on lidar data for autonomous vehicle applications.
\newblock In {\em IEEE International Conference on Robotics and Automation}, pages 5067--5073, 2017.

\bibitem{zhou2020polarNet}
Yang Zhang, Zixiang Zhou, Philip David, Xiangyu Yue, Zerong Xi, Boqing Gong, and Hassan Foroosh.
\newblock Polarnet: An improved grid representation for online lidar point clouds semantic segmentation.
\newblock In {\em IEEE/CVF Conference on Computer Vision and Pattern Recognition}, pages 9601--9610, 2020.

\bibitem{zhao2021fidnet}
Yiming Zhao, Lin Bai, and Xinming Huang.
\newblock Fidnet: Lidar point cloud semantic segmentation with fully interpolation decoding.
\newblock In {\em IEEE/RSJ International Conference on Intelligent Robots and Systems}, pages 4453--4458, 2021.

\bibitem{zhu2021cylindrical}
Xinge Zhu, Hui Zhou, Tai Wang, Fangzhou Hong, Yuexin Ma, Wei Li, Hongsheng Li, and Dahua Lin.
\newblock Cylindrical and asymmetrical 3d convolution networks for lidar segmentation.
\newblock In {\em IEEE/CVF Conference on Computer Vision and Pattern Recognition}, pages 9939--9948, 2021.

\bibitem{CBST}
Yang Zou, Zhiding Yu, B~V K~Vijaya Kumar, and Jinsong Wang.
\newblock Unsupervised domain adaptation for semantic segmentation via class-balanced self-training.
\newblock In {\em European Conference on Computer Vision}, pages 289--305, 2018.

\end{thebibliography}
}

\end{document}